\documentclass[journal]{IEEEtran}
\IEEEoverridecommandlockouts
\usepackage{amsmath,amssymb,amsthm}
\usepackage{graphicx}
\usepackage{booktabs}
\usepackage{cite}
\usepackage{xcolor}
\usepackage{algorithm}
\usepackage{algpseudocode}
\usepackage{tikz}
\usetikzlibrary{positioning,arrows.meta}
\usepackage[hidelinks]{hyperref}
\newtheorem{theorem}{Theorem}
\newtheorem{proposition}{Proposition}
\newtheorem{lemma}{Lemma}
\newtheorem{assumption}{Assumption}
\theoremstyle{remark}
\newtheorem{remark}{Remark}
\newcommand{\R}{\mathbb{R}}
\newcommand{\SO}{\mathrm{SO}(3)}
\newcommand{\inprogress}[1]{}
\begin{document}
\title{Convergence of Hybrid Visual Servoing as a Gradient Flow: Weight Thresholds and Certified Regions of Attraction}
\author{A.~H.~Abdul~Hafez%
\thanks{This work has been submitted to the IEEE for possible publication. Copyright may be transferred without notice, after which this version may no longer be accessible.}%
\thanks{A.~H.~Abdul~Hafez is with King Faisal University, Al-Ahsa, Saudi Arabia (e-mail: aabdulhafiz@kfu.edu.sa).}}
\maketitle

\begin{abstract}
Hybrid visual servoing adds a weighted pose error to the image error and avoids image-based local minima in practice, yet no guarantee states from where it converges. With a smooth rotation error, the hybrid law is a gradient flow for every weight, with a nondegenerate goal that, for topological reasons, cannot be globally attractive. In five benchmark geometries, every spurious critical point found above the fold of the image-based minima is a saddle near $180^\circ$. We prove an explicit weight threshold for any geometry and certify the folds by interval arithmetic; of 496 random geometries, only six had an identified fold above $0.1$. Below the level needed to approach an object point, we prove convergence from sublevel sets free of spurious critical points, and a six-dimensional computer-assisted proof certifies such sets for all five geometries; at unit weight their certified pure-rotation radii range from $13.93^\circ$ to $35.61^\circ$. A barrier in the potential guarantees visibility, and convergence under a numerically verified hypothesis. Of 2600 simulated robot-in-the-loop runs of the smooth hybrid law, 2438 converged without feature loss, ending within 1.76\,mm and $0.48^\circ$ of the goal, and 162 reached a joint limit; position-based servoing lost features in 24 runs.
\end{abstract}


\section{Introduction}\label{sec:intro}
Visual servoing closes a control loop around a camera~\cite{chaumette2006visual,chaumette2007visual}. Image-based visual servoing (IBVS) regulates image features directly and is robust to calibration errors, but its cost can have local minima, and the camera can retreat or rotate in ways that leave the workspace~\cite{chaumette1998potential}. Position-based visual servoing (PBVS) regulates the estimated pose, has a unique minimum and a straight camera path, but loses control of the image, so features can leave the field of view. Hybrid schemes combine the two errors, either in partitioned laws~\cite{malis1999,corke2001}, by switching~\cite{gans2007}, or as a weighted sum of the two costs~\cite{hafez2006,hafez2007,hafez2013,hafez2014,kermorgant2011visibility,kermorgant2014}. In practice, hybrid laws avoid the local minima of IBVS and keep the features in view better than PBVS.

Why they do, from which weight on, and from which initial poses, has remained open. Arguments for the absence of local minima~\cite[Sec.~IV]{hafez2007} required each error block to lie in the kernel of its own transposed Jacobian at an equilibrium, while an equilibrium only requires the weighted sum of the two gradient terms to vanish. Recent work computed all equilibria of IBVS for benchmark geometries~\cite{colotti2024,colotti2023thesis} and showed that the local minima are common; recent work~\cite{hafez2026ral} showed that, for the hybrid law with the $\theta u$ rotation error, each IBVS minimum is destroyed in a saddle-node fold once the pose weight exceeds a small threshold. That letter gave computed thresholds for five geometries, not guarantees, and its rotation error has a cut locus at $180^\circ$ through which critical points leave the domain.

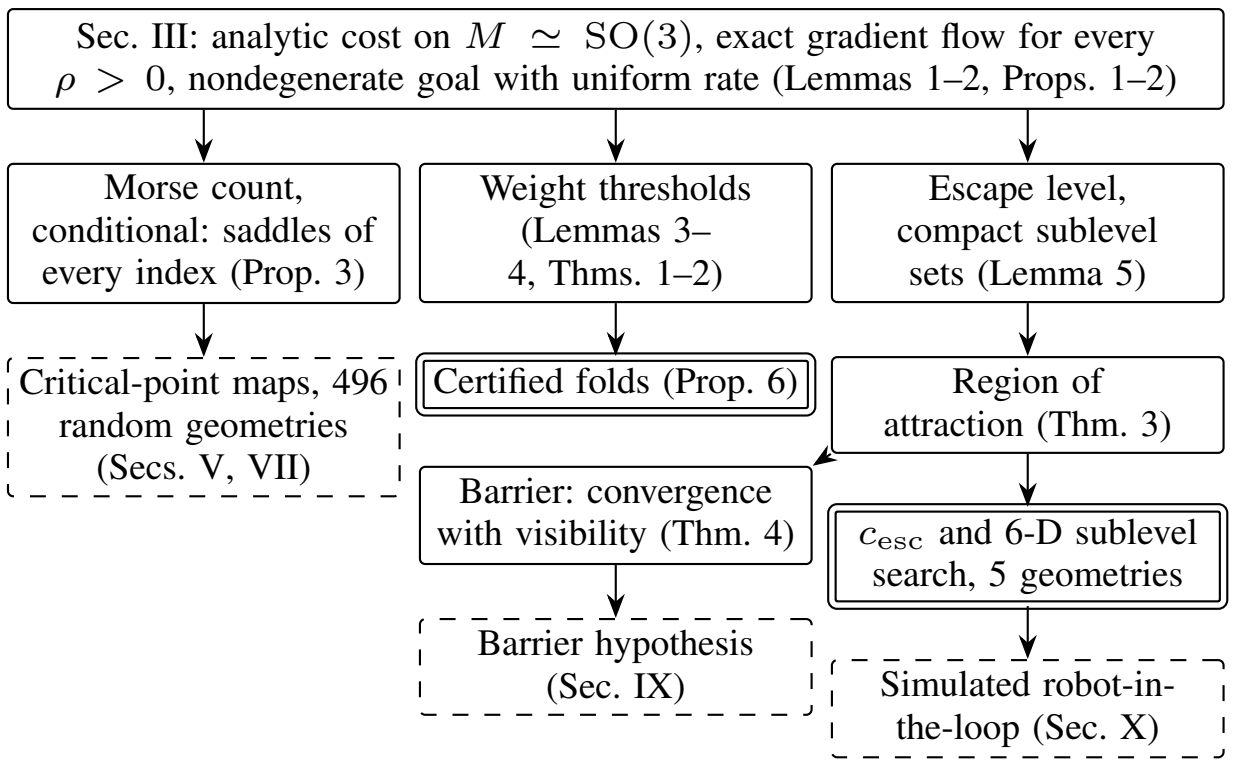
\begin{figure}[t]
\centering
\begin{tikzpicture}[font=\scriptsize, node distance=3.5mm and 2mm,
  pr/.style={draw, rounded corners=1pt, align=center, text width=#1, inner sep=2pt},
  pr/.default=2.45cm,
  ca/.style={pr=#1, double, double distance=1pt}, ca/.default=2.45cm,
  nu/.style={pr=#1, dashed}, nu/.default=2.45cm,
  ar/.style={-{Stealth[length=1.6mm]}}]
\node[pr=7.9cm] (s3) {Sec.~\ref{sec:formulation}: analytic cost on $M\simeq\SO$, exact gradient flow for every $\rho>0$, nondegenerate goal with uniform rate (Lemmas~\ref{lem:domain}--\ref{lem:rot}, Props.~\ref{prop:flow}--\ref{prop:goal})};
\node[pr, below=of s3.south west, anchor=north west] (mo) {Morse count, conditional: saddles of every index (Prop.~\ref{prop:morse})};
\node[pr, below=of s3.south, anchor=north] (th) {Weight thresholds (Lemmas~\ref{lem:weight}--\ref{lem:descent}, Thms.~\ref{thm:large}--\ref{thm:cert})};
\node[pr, below=of s3.south east, anchor=north east] (cp) {Escape level, compact sublevel sets (Lemma~\ref{lem:compact})};
\node[nu, below=of mo] (mp) {Critical-point maps, 496 random geometries (Secs.~\ref{sec:maps}, \ref{sec:random})};
\node[ca, below=of th] (fo) {Certified folds (Prop.~\ref{prop:fold})};
\node[pr, below=of cp] (ro) {Region of attraction (Thm.~\ref{thm:roa})};
\node[ca, below=of ro] (rc) {$c_{\rm esc}$ and 6-D sublevel search, 5 geometries};
\node[pr, below=of fo] (ba) {Barrier: convergence with visibility (Thm.~\ref{thm:barrier})};
\node[nu, below=of ba] (bh) {Barrier hypothesis (Sec.~\ref{sec:barrier})};
\node[nu, below=of rc] (si) {Simulated robot-in-the-loop (Sec.~\ref{sec:experiments})};
\draw[ar] (s3.south west -| mo.north) -- (mo); \draw[ar] (s3.south -| th.north) -- (th); \draw[ar] (s3.south -| cp.north) -- (cp);
\draw[ar] (th) -- (fo); \draw[ar] (cp) -- (ro); \draw[ar] (ro) -- (rc); \draw[ar] (mo) -- (mp);
\draw[ar] (ro.south west) -- (ba.north east); \draw[ar] (ba) -- (bh); \draw[ar] (rc) -- (si);
\end{tikzpicture}
\caption{Structure of the results. Solid boxes: proved for any geometry. Double boxes: computer-assisted proofs (interval arithmetic) for specific geometries. Dashed boxes: numerical evidence and simulation.}
\label{fig:roadmap}
\end{figure}

This paper turns those observations into certified guarantees. We use a smooth rotation error, so that the closed loop is a gradient flow of a smooth cost on the space of camera poses with every object point in front of the camera, and we combine three tools: Morse theory for the global structure of the critical points, explicit inequalities for a weight threshold valid in any geometry, and interval arithmetic~\cite{moore2009,tucker2011} for computer-assisted proofs of the remaining hypotheses in concrete geometries. The contributions are:
\begin{enumerate}
\item \emph{Gradient-flow structure} (Section~\ref{sec:formulation}). With the smooth rotation error, the cost is real-analytic on the configuration space, the law is a gradient flow of the cost for a metric that is positive definite for every weight, the goal is nondegenerate with a uniform local decay rate $\lambda$ in every direction, and the configuration space has the topology of $\SO$.
\item \emph{Critical-point structure} (Sections~\ref{sec:topology}--\ref{sec:maps}). The configuration space is not contractible, so no continuous law makes the goal globally attractive. Under a properness assumption, which fails on the full domain of our benchmarks, Morse theory forces at least one saddle of each index 1, 2, 3 for every weight. Numerical maps of five benchmark geometries at weights from $10^{-4}$ to $10$ show that above the fold every spurious critical point found is a saddle near $180^\circ$.
\item \emph{Weight thresholds} (Section~\ref{sec:certificate}). An explicit threshold above which the goal is the only local minimum in a visible region, with constants in closed form for any geometry; certified saddle-node folds, with enclosures of ten digits, for the four IBVS minima of the benchmarks; and a study of 496 random geometries (Section~\ref{sec:random}) in which every spurious minimum found is an IBVS minimum ending in a fold, and the fold exceeds $0.1$ in 1.2\% of the geometries.
\item \emph{Certified regions of attraction} (Section~\ref{sec:roa}). We show that the flow can leave the domain through an object point, compute the level $c_{\rm esc}$ at which this becomes possible, and prove that every trajectory starting below a level that contains no spurious critical point converges. A six-dimensional computer-assisted proof checks the hypothesis for all five geometries. A cone condition on the realised twist extends the convergence to estimated features, depths and pose, and, checked by sampling, holds on every certified set for the calibration errors of our experiments.
\item \emph{Guaranteed visibility} (Section~\ref{sec:barrier}). A flat-top barrier inside the potential keeps every feature in the image and every depth above a floor, with the same convergence theorem and without an escape level. Its hypothesis is verified numerically, and certified where the barrier is inactive.
\item \emph{Simulated robot-in-the-loop validation} (Section~\ref{sec:experiments}). 9600 runs with a simulated 7-DOF arm, rendered images, image processing, pose estimation, latency and calibration errors compare the hybrid laws with IBVS, PBVS, 2.5D visual servoing~\cite{malis1999} and the scheme of~\cite{kermorgant2014}, from the certified regions and from wide start sets; position-based servoing lost features in 24 runs, the hybrid laws in none.
\end{enumerate}
Fig.~\ref{fig:roadmap} shows how the results depend on each other and which are proved, certified by computer or observed. Section~\ref{sec:discussion} discusses what the guarantees do and do not cover, and Section~\ref{sec:conclusion} concludes. Code, certificate logs and data will be released with the paper.

\section{Related Work}\label{sec:related}
\paragraph*{Equilibria of IBVS} The local minima of image-point IBVS were exhibited early~\cite{chaumette1998potential}. Colotti et al.~\cite{colotti2024} computed all stable and unstable equilibria of IBVS for benchmark geometries with certified numerical methods, and showed for a cylinder target that global stability can be obtained by a suitable choice of features~\cite{colotti2025cylinder}; their geometries, together with the singular configurations of the perspective-four-point problem~\cite{pascual2021}, are our benchmarks. Stability of 3-D feature-based servoing was analysed in~\cite{costanzo2022}, and the stability and regions of attraction of IBVS in~\cite{colotti2023thesis}; stability boundaries of general nonlinear systems are characterised in~\cite{chiang1988}.

\paragraph*{Hybrid and partitioned laws} 2.5D visual servoing~\cite{malis1999,malis2002} decouples rotation and translation using a partial pose estimate; partitioned IBVS~\cite{corke2001} treats the optical-axis motion separately; switched schemes~\cite{gans2007} alternate between IBVS and PBVS with a Lyapunov argument. Weighted combinations of the image and pose costs were introduced in~\cite{hafez2006,hafez2007} and developed with visibility and path terms in~\cite{hafez2008,hafez2013,hafez2014}. Kermorgant and Chaumette combine PBVS with image features whose weights grow near the image border~\cite{kermorgant2011visibility} and generalise the idea to joint limits and other constraints~\cite{kermorgant2011joint,kermorgant2014}. Recent work~\cite{hafez2026ral} explains, through saddle-node folds, why a small pose term removes the IBVS minima; this paper gives the corresponding guarantees.

\paragraph*{Visibility} Path planning in the image keeps features in view and then tracks the planned trajectory with IBVS~\cite{mezouar2002}. Switching among position-based strategies and backward motion keeps all features in view in the absence of extrinsic errors~\cite{chesi2004}. Navigation functions on a model of the visible set give convergence from almost every visible configuration while keeping all features in view~\cite{cowan2002}, building on~\cite{rimon1992}. Predictive control handles visibility as an output constraint~\cite{allibert2010}, and control barrier functions~\cite{ames2017} make constraint sets forward invariant in a quadratic program. Related work~\cite{hafez2026whvs} certifies visibility between sampling instants for prescribed weight schedules. Our barrier is part of the potential rather than a constraint on the input, so that the convergence and the visibility proofs are one argument.

\paragraph*{Topology and computer-assisted proofs} No continuous feedback can globally stabilise a point of $\SO$~\cite{bhat2000}; Morse theory~\cite{milnor1963} makes the obstruction quantitative for gradient flows, as navigation functions do for obstacle spaces~\cite{rimon1992}. Interval arithmetic gives rigorous enclosures of functions over boxes~\cite{moore2009,tucker2011} and has been used to certify the equilibria of IBVS~\cite{colotti2024}; we use it for folds, for an escape level and for six-dimensional searches over sublevel sets.

\section{Smooth Hybrid Cost and Its Gradient Flow}\label{sec:formulation}
This section defines the object that the rest of the paper analyses and proves the properties on which every later result rests. The cost is real-analytic on the whole configuration space. The law is a gradient flow of the cost for a metric that is positive definite for every weight. The goal is a nondegenerate minimum at which every direction decays at the same rate. And the configuration space has the topology of $\SO$. The first property fails for the $\theta u$ error of~\cite{hafez2026ral} at $180^\circ$, and the earlier analyses of weighted hybrid laws~\cite{hafez2006,hafez2007} did not establish the others.

\subsection{Configuration space and features}
The camera pose $(T,R)$ is expressed in the desired camera frame, and the camera twist $\xi=(v,\omega)$ in the current camera frame, so that $\dot T=Rv$ and $\dot R=R[\omega]_\times$. The camera observes $n\ge1$ points $P_i$, with camera coordinates $(X_i,Y_i,Z_i)^\top=R^\top(P_i-T)$ and normalised image coordinates $s_i=(x_i,y_i)=(X_i/Z_i,Y_i/Z_i)$. Then $\dot s_i=L_i\xi$ with the interaction matrix
\begin{equation}\label{eq:L}
L_i=\begin{bmatrix}-1/Z_i&0&x_i/Z_i&x_iy_i&-(1+x_i^2)&y_i\\0&-1/Z_i&y_i/Z_i&1+y_i^2&-x_iy_i&-x_i\end{bmatrix},
\end{equation}
and $L$ stacks the $L_i$. The configuration space is the set of poses with every point in front of the camera,
\[
M=\{(T,R)\in\R^3\times\SO:\ Z_i>0,\ i=1,\dots,n\}.
\]
\begin{lemma}[Configuration space]\label{lem:domain}
$M$ is an open subset of $\R^3\times\SO$ that deformation-retracts onto a copy of $\SO$. Hence $M$ is homotopy equivalent to $\SO\cong\mathbb{RP}^3$.
\end{lemma}
\begin{proof}
Let $r_3$ be the third column of $R$. Then $Z_i=r_3^\top(P_i-T)$, so $(T,R)\in M$ if and only if $r_3^\top T<m(R)=\min_ir_3^\top P_i$. For fixed $R$ the fibre is an open half-space, and $m$ is continuous. The map $\sigma(R)=\big((m(R)-\bar Z)r_3,R\big)$ is a continuous section of $M$, since $r_3^\top(m(R)-\bar Z)r_3=m(R)-\bar Z<m(R)$. The homotopy $H_t(T,R)=\big((1-t)T+t(m(R)-\bar Z)r_3,\,R\big)$ stays in $M$ by convexity of the fibres, fixes the section, and ends on it.
\end{proof}
With $e=s-s^*$ and the mean desired depth $\bar Z$, the smooth hybrid cost is
\begin{equation}\label{eq:cost}
E_\rho(T,R)=\tfrac12\|e\|^2+\rho\,P(T,R),\qquad P(T,R)=\frac{\|T\|^2}{2\bar Z^2}+1-\cos\theta,
\end{equation}
where $\theta$ is the rotation angle of $R$. The cost is invariant under the scaling $(P_i,T,\bar Z)\mapsto k(P_i,T,\bar Z)$, $k>0$, so the weight $\rho$ is dimensionless and every result below holds for all scaled copies of a geometry. We use $T/\bar Z$ throughout.

\subsection{The smooth rotation error}
The rotation term is the chordal distance of attitude control~\cite{koditschek1989,chaturvedi2011}, $1-\cos\theta=\tfrac12(3-\operatorname{tr}R)=\tfrac14\|R-I\|_F^2$. Its new role here is as the pose term of a hybrid visual-servoing law. For any function $f$ on $\SO$ we take the body gradient and Hessian along the curves $t\mapsto R\,\mathrm{Exp}(t\omega)$.
\begin{lemma}[Rotation term]\label{lem:rot}
Let $f(R)=1-\cos\theta$ and $R=\mathrm{Exp}(\theta u)$. Then:
\begin{enumerate}
\item $f$ is a polynomial in the entries of $R$, hence real-analytic on $\SO$;
\item its body gradient is $\nabla f(R)=\operatorname{vee}\big(\tfrac12(R-R^\top)\big)=\sin\theta\,u$, a polynomial in $R$;
\item its body Hessian is $\tfrac12\big[(1+\cos\theta)I-(1-\cos\theta)uu^\top\big]$, with eigenvalue $\tfrac12(1+\cos\theta)$ twice, orthogonal to $u$, and $\cos\theta$ along $u$;
\item its critical points are $R=I$, a nondegenerate minimum with Hessian $I$, and the half-turns $\theta=\pi$, a critical manifold diffeomorphic to $\mathbb{RP}^2$ on which the Hessian has eigenvalues $0,0,-1$.
\end{enumerate}
\end{lemma}
\begin{proof}
(i) is the trace formula. For (ii), $\frac{d}{dt}f(R\,\mathrm{Exp}(t\omega))|_{t=0}=-\tfrac12\operatorname{tr}(R[\omega]_\times)$. Only the skew part $\tfrac12(R-R^\top)=\sin\theta[u]_\times$ of $R$ contributes, and $\operatorname{tr}([a]_\times[\omega]_\times)=-2a^\top\omega$. (iii) is computed in Appendix~A. (iv) follows from (ii) and (iii): $\sin\theta\,u=0$ exactly when $\theta\in\{0,\pi\}$, and the half-turns $\mathrm{Exp}(\pi u)$, $u\in S^2/\{\pm1\}$, form a copy of $\mathbb{RP}^2$ whose tangent directions are the $\omega\perp u$.
\end{proof}
For comparison, the $\theta u$ error of~\cite{hafez2026ral} uses $\tfrac12\theta^2$, whose body gradient $\theta u$ jumps from $\pi u$ to $-\pi u$ across the half-turns. That cost is continuous but not differentiable there, its law is discontinuous, and critical points can disappear into that set, as observed in~\cite{hafez2026ral}. Near the goal the two errors agree to second order, $1-\cos\theta=\tfrac12\theta^2+O(\theta^4)$, so the two laws have the same linearisation there (Proposition~\ref{prop:goal}).

\begin{remark}[Where the far saddles come from]\label{rem:bott}
By Lemma~\ref{lem:rot}(iv), the pose term $P$ is a Morse--Bott function. Its critical points are the goal and the half-turns at $T=0$, a degenerate family with one direction of negative curvature (along the axis) and two flat directions. Writing $E_\rho=\rho\,(P+\rho^{-1}\tfrac12\|e\|^2)$, the image term is a perturbation that breaks the degeneracy of this family wherever it meets $M$, that is, for the half-turns whose axes keep every point in front of the camera, close to the optical axis. The perturbation leaves isolated critical points with the negative axis direction and with indices from 1 to 3. This is the mechanism behind the spurious saddles near $180^\circ$ found in Section~\ref{sec:maps}. The remark is a heuristic, because the family meets $M$ in a noncompact set. Under Assumption~\ref{as:proper}, Proposition~\ref{prop:morse} would make the existence of saddles of each index a theorem; the assumption fails on $M$ for our benchmarks (Section~\ref{sec:topology}), so the saddles found there are numerical observations.
\end{remark}

\subsection{The hybrid law is a gradient flow}
Let $D=\operatorname{diag}(\bar Z^{-2}I_3,I_3)$ and, with tildes, use the scaled translation $\bar Z v$, so that $\tilde D=I$.
\begin{proposition}[Gradient flow]\label{prop:flow}
Let $\rho>0$. On $M$:
\begin{enumerate}
\item the body gradient of $E_\rho$ is
\begin{equation}\label{eq:grad}
g=f_0+\rho\,p,\qquad f_0=L^\top e,\qquad p=\Big(\frac{R^\top T}{\bar Z^2},\ \sin\theta\,u\Big);
\end{equation}
\item the matrix $G=L^\top L+\rho D$ is positive definite whatever the rank of $L$, with $\rho\le\lambda_{\min}(\tilde G)\le\lambda_{\max}(\tilde G)\le\|\tilde L\|^2+\rho$;
\item the law $\xi=-\lambda G^{-1}g$, $\lambda>0$, is real-analytic on $M$, and along its trajectories
\begin{equation}\label{eq:Edot}
\dot E_\rho=-\lambda\,g^\top G^{-1}g\le-\frac{\lambda\,\|\tilde g\|^2}{\|\tilde L\|^2+\rho};
\end{equation}
\item it is the gradient flow of $E_\rho$ for the Riemannian metric $\langle\xi,\eta\rangle_x=\xi^\top G(x)\eta$, so its equilibria are exactly the critical points of $E_\rho$. These are the poses at which the image force balances the pose force, $L^\top e=-\rho\,p$.
\end{enumerate}
\end{proposition}
\begin{proof}
(i) $\frac{d}{dt}\tfrac12\|e\|^2=e^\top L\xi$, $\frac{d}{dt}\|T\|^2/(2\bar Z^2)=T^\top Rv/\bar Z^2$, and Lemma~\ref{lem:rot}(ii) gives the rotation part. (ii) $L^\top L\succeq0$ and $\tilde D=I$; the upper bound is the triangle inequality. (iii) Every entry of $L$, $e$ and $p$ is rational in the pose with denominators $Z_i>0$, and $G^{-1}$ exists by (ii). Then $\dot E_\rho=g^\top\xi$, and the bound follows from (ii). (iv) The gradient of $E_\rho$ for the metric $G$ is the vector $\eta$ with $\eta^\top G\zeta=g^\top\zeta$ for all $\zeta$, that is $\eta=G^{-1}g$. Since $G\succ0$, $\dot E_\rho=0$ only where $g=0$.
\end{proof}
Item (iv) is the exact equilibrium condition. It requires only the weighted sum of the two forces to vanish, not each block separately, and this is why the sufficient condition used in~\cite[Sec.~IV]{hafez2007} does not exclude spurious equilibria. Item (ii) is where the pose term matters beyond the cost: for IBVS ($\rho=0$) the matrix $L^\top L$ is singular wherever $L$ loses rank, and the law is then not a gradient flow. For every $\rho>0$ the Lyapunov identity~\eqref{eq:Edot} holds on all of $M$, with no exceptional set, so the cost itself is a Lyapunov function and its sublevel sets are forward invariant (Sections~\ref{sec:roa} and~\ref{sec:barrier}).

\subsection{The goal}
\begin{proposition}[Local behaviour at the goal]\label{prop:goal}
For every $\rho>0$ and every geometry, the goal $x^*=(0,I)$ is a nondegenerate local minimum of $E_\rho$ with Hessian
\[
H(x^*)=L^{*\top}L^*+\rho D=G(x^*),
\]
where $L^*=L(x^*)$. Consequently, the linearisation of the law at $x^*$ in body coordinates is $\dot\zeta=-\lambda\zeta$: every direction of pose error decays at the rate $\lambda$, independent of the weight, the geometry and the rank of $L^*$.
\end{proposition}
\begin{proof}
The Hessian of $\tfrac12\|e\|^2$ is $L^\top L+\sum_ke_k\nabla^2s_k$, and $e=0$ at $x^*$. The pose term has Hessian $\bar Z^{-2}I$ in translation and, by Lemma~\ref{lem:rot}(iv), $I$ in rotation. So $H(x^*)=G(x^*)\succ0$ by Proposition~\ref{prop:flow}(ii). Near $x^*$, $g=H(x^*)\zeta+O(\|\zeta\|^2)$, so $\dot\zeta=-\lambda G(x^*)^{-1}H(x^*)\zeta+O(\|\zeta\|^2)=-\lambda\zeta+O(\|\zeta\|^2)$.
\end{proof}
The Gauss--Newton law for IBVS has the same property only at goals where $L^*$ has full rank; the pose term makes it hold for every geometry, with $n\ge1$ points. The nondegeneracy of $x^*$ is also what the computer-assisted proofs of Sections~\ref{sec:certificate} and~\ref{sec:roa} use: the Krawczyk test certifies the goal as the only zero of $g$ in a box around it because $D_yg(x^*)$ is invertible.

\subsection{A rational chart for computation}
We parametrise rotations by the modified Rodrigues parameters $\sigma=\tan(\theta/4)\,u$~\cite{shuster1993}, with
\[
R(\sigma)=I+\frac{8[\sigma]_\times^2+4(1-\|\sigma\|^2)[\sigma]_\times}{(1+\|\sigma\|^2)^2}.
\]
The closed ball $\|\sigma\|\le1$ covers $\SO$, and its boundary sphere is the set of half-turns. In the coordinates $y=(T/\bar Z,\sigma)$ the features, $L$, $e$, $\cos\theta=\tfrac12(\operatorname{tr}R-1)$ and $\sin\theta\,u$ (Lemma~\ref{lem:rot}(ii)) are rational functions whose only denominators are $(1+\|\sigma\|^2)^2$ and the depths $Z_i$. Hence $E_\rho$, $g$ and $G$ are rational in $y$ on $M$, regular across $\theta=\pi$, and can be enclosed over a box by interval arithmetic using $+,-,\times,\div$ only. The $\theta u$ error has no such chart near $\theta=\pi$.

\subsection{What the formulation provides}
Each property above enables one later step. Analyticity on $M$ (Lemma~\ref{lem:rot}) and the topology of $M$ (Lemma~\ref{lem:domain}) allow the Morse count of Section~\ref{sec:topology}. The exact equilibrium condition (Proposition~\ref{prop:flow}(iv)) underlies the critical-weight and threshold results of Section~\ref{sec:certificate}. The Lyapunov identity with a metric bounded below, on all of $M$ and for every weight (Proposition~\ref{prop:flow}(ii)--(iii)), gives the forward invariance used in Sections~\ref{sec:roa} and~\ref{sec:barrier}. The nondegenerate goal (Proposition~\ref{prop:goal}) gives the goal box of the certificates. And the rational chart makes all of them checkable by interval arithmetic.

\section{Topology of the Critical Points}\label{sec:topology}
\begin{assumption}\label{as:proper}
For the weight $\rho>0$ considered, $E_\rho$ is a Morse function on $M$ and its sublevel sets are compact.
\end{assumption}
Compactness holds as $\|T\|\to\infty$ because of the pose term, and as $Z_i\to0^+$ with $(X_i,Y_i)$ bounded away from zero because $\|s_i\|\to\infty$. Compactness fails only when the camera centre approaches an object point $P_i$: then $Z_i\to0$ while $s_i$ can stay bounded. This case does occur. At large weights the far saddles converge onto an object point (in Case 1a their distance to $P_3$ falls from $0.46\bar Z$ at $\rho=1$ to $0.21\bar Z$ at $\rho=10$), because the image gradient of a nearby point grows like the inverse distance and can balance any pose force. Assumption~\ref{as:proper} therefore does not hold on all of $M$, and Proposition~\ref{prop:morse} is a conditional result. The theorems of Section~\ref{sec:certificate} do not use it; the saddles near object points have a depth below $0.2\bar Z$ and lie outside the visible regions $K$ used there.

\begin{proposition}[Morse count]\label{prop:morse}
Under Assumption~\ref{as:proper}, let $n_k$ be the number of critical points of $E_\rho$ of index $k$. Then $\sum_k(-1)^k n_k=\chi(\SO)=0$, and $n_k\ge1$ for $k=0,1,2,3$. Since the goal is a critical point of index $0$, there is at least one saddle of each index 1, 2 and 3, and the spurious critical points satisfy $\sum_{\text{spurious}}(-1)^{\mathrm{index}}=-1$.
\end{proposition}
\begin{proof}
By Lemma~\ref{lem:domain}, $M$ is homotopy equivalent to $\SO\cong\mathbb{RP}^3$, whose Betti numbers with $\mathbb{Z}_2$ coefficients are $1,1,1,1$ and whose Euler characteristic is $0$. For a proper Morse function bounded below, $M$ is homotopy equivalent to a CW complex with one cell of dimension $k$ per critical point of index $k$~\cite{milnor1963}. The weak Morse inequalities give $n_k\ge b_k=1$, and the Euler characteristic gives the alternating sum.
\end{proof}

Two consequences matter for control. Independently of the assumption, $M$ is not contractible (Lemma~\ref{lem:domain}), so no continuous law, this one included, makes the goal globally asymptotically stable on $M$~\cite{bhat2000}. Where Assumption~\ref{as:proper} holds, no choice of weight removes the saddles, and the count gives a test for numerical searches: a computed set of critical points with a different alternating sum, or a missing index, is incomplete. On our benchmarks the assumption fails, and the saddles near $180^\circ$ found at every weight of the maps (Section~\ref{sec:maps}) are observations, not consequences of Proposition~\ref{prop:morse}.

\section{Critical-Point Maps}\label{sec:maps}
We searched for all critical points of~\eqref{eq:cost} for Colotti Cases 1a, 1b, 2a, 2b~\cite{colotti2024} and Pascual-Escudero Example~1~\cite{pascual2021} at 11 weights from $10^{-4}$ to $10$. Each set combines a batched Levenberg--Marquardt search from 40\,000 starts, a scalar multi-start search, continuation of every branch in $\rho$ in both directions, and dense local searches near near-degenerate points.

\begin{table}[t]
\centering
\caption{Fold of each IBVS local minimum for the smooth rotation error. The values are certified to the digits shown (Section~\ref{sec:certificate}).}
\label{tab:folds}
\begin{tabular}{lc}
\toprule
Geometry & Fold $\rho$ \\
\midrule
Colotti 1a & $8.1727\times10^{-4}$ \\
Colotti 1b & $1.0893\times10^{-3}$ \\
Colotti 2a & $2.2506\times10^{-4}$ \\
Colotti 2b & no minimum \\
P.-E. Ex.~1 & $1.9175\times10^{-5}$ \\
\bottomrule
\end{tabular}
\end{table}

\begin{figure*}[t]
\centering
\includegraphics[width=\textwidth]{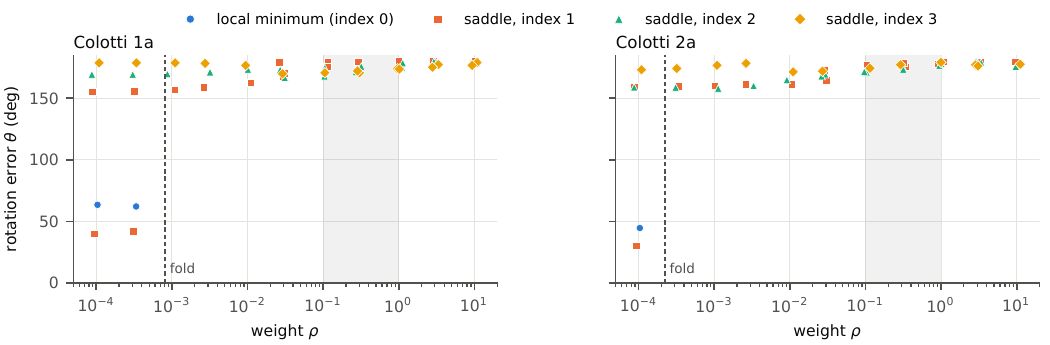}
\caption{Spurious critical points of the smooth hybrid cost against the weight, by index (points jittered slightly in $\rho$). Dashed: the fold of the IBVS minimum. Shaded: the design range $0.1\le\rho\le1$.}
\label{fig:crit}
\end{figure*}

Table~\ref{tab:folds} lists the folds; they lie between $2\times10^{-5}$ and $1.1\times10^{-3}$, close to those reported for the $\theta u$ error in~\cite{hafez2026ral}. Fig.~\ref{fig:crit} shows the critical points of Cases 1a and 2a. The only local minima other than the goal are the IBVS minima below their folds. Above the fold, every spurious critical point in the five geometries is a saddle with a rotation error above $140^\circ$, and in the design range $0.1\le\rho\le1$ of~\cite{hafez2026ral} all of them lie within $13^\circ$ of $180^\circ$. Unlike with the $\theta u$ error, far saddles were found at every sampled weight.

In 52 of the 55 maps, the observed index sum agrees with the conditional Morse prediction of Proposition~\ref{prop:morse}. The remaining three, at $\rho=3$ in Cases 1a, 1b and 2a, have a discrepant sum, and their completeness is unresolved; at these weights several saddles lie within $0.1$ to $0.4\bar Z$ of an object point, where the cost is nearly singular and the search is least reliable. In the five geometries, no map contains a local minimum other than the goal at a weight above its fold.

\section{Threshold Certificate}\label{sec:certificate}
\subsection{Pointwise conditions}
\begin{lemma}[Critical weight]\label{lem:weight}
Let $p(x)\ne0$. The point $x$ is critical for some $\rho>0$ only if $f_0(x)$ is antiparallel to $p(x)$, and then only for $\rho=\hat\rho(x)=-\langle f_0,p\rangle/\|p\|^2$.
\end{lemma}
\begin{proof}
$f_0+\rho p=0$ forces $f_0=-\rho p$; taking the inner product with $p$ gives $\hat\rho$.
\end{proof}

\begin{lemma}[Descent direction]\label{lem:descent}
Let $\theta(x)>\pi/2$ and $h(x)=\frac{d^2}{dt^2}\tfrac12\|e(T,R\,\mathrm{Exp}(tu))\|^2\big|_{t=0}$. If $x$ is critical for $\rho>h(x)^+/(-\cos\theta)$, then $x$ is not a local minimum.
\end{lemma}
\begin{proof}
Along $t\mapsto(T,R\,\mathrm{Exp}(tu))$ the rotation angle is $\theta+t$, so the second derivative of $E_\rho$ is $h+\rho\cos\theta<0$. At a critical point this second derivative is the Hessian in the direction $(0,u)$.
\end{proof}
More generally, by Lemma~\ref{lem:rot}(iii) the only negative eigenvalue of the pose Hessian, $\cos\theta$, lies along the rotation axis.

\begin{proposition}[Pointwise threshold]\label{prop:pointwise}
Let $N_0$ be a geodesically convex neighbourhood of the goal on which the Hessians of $\tfrac12\|e\|^2$ and of the pose term are positive definite, so that for every $\rho>0$ the goal is the only critical point of $E_\rho$ in $N_0$. For $x\in K\setminus N_0$ let $d(x)=h(x)^+/(-\cos\theta(x))$ if $\theta(x)>\pi/2$ and $d(x)=+\infty$ otherwise, and let
\[
\tau(x)=\begin{cases}\min\{\hat\rho(x),d(x)\} & \text{in case (a)},\\ d(x) & \text{in case (b)},\\ 0 & \text{otherwise,}\end{cases}
\]
where (a) means $p(x)\ne0$, $\hat\rho(x)>0$ and $f_0(x)=-\hat\rho(x)p(x)$, and (b) means $p(x)=0$ and $f_0(x)=0$.
If $\rho>\sup_{K\setminus N_0}\tau$, no critical point of $E_\rho$ in $K$ other than the goal is a local minimum.
\end{proposition}
\begin{proof}
Otherwise $x$ is critical for no $\rho>0$ (Lemma~\ref{lem:weight}; if $p(x)=0$ and $f_0(x)\ne0$, then $g=f_0\ne0$). In case (b), which by $p=0$ consists of half-turns at $T=0$, $x$ is critical for every $\rho$, and $\rho>d(x)$ makes it a non-minimum by Lemma~\ref{lem:descent}. In case (a) $x$ is critical only for $\rho=\hat\rho(x)$; then $\rho>\tau(x)$ rules out $\tau(x)=\hat\rho(x)$, so $\tau(x)=d(x)<\rho$, and Lemma~\ref{lem:descent} applies. In $N_0$ the goal is the only critical point.
\end{proof}

The fold bounds any such threshold from below: if the IBVS minimum lies in $K$, every valid threshold, $\sup_{K\setminus N_0}\tau$ included, is at least $\rho_f$. Evaluating $\sup\tau$ requires knowing where case (a) holds, that is, locating the candidate critical points, which is what the box-wise certificate below does. As an indication only, we sampled the relaxation $\tilde\tau=\min\{\hat\rho^+,d\}$, which drops the antiparallel condition and is at least $\tau$ wherever $p\ne0$. For Case 1a, with $K$ restricted to camera positions within $3\bar Z$, depths of at least $0.1$ to $0.5\bar Z$ and points within $1.3$ times the largest desired image coordinate, its sampled maximum is about $8.5$ to $13$. With a depth floor of $0.25\bar Z$ and the full-Hessian version of Lemma~\ref{lem:descent}, it is about $5$. For Case 1a, the sampled relaxation takes values far above the identified fold weight. The sharpness of the exact threshold in Proposition~\ref{prop:pointwise} remains open.

\subsection{Explicit large-weight theorem}
Lemmas~\ref{lem:weight} and~\ref{lem:descent} give a threshold that holds for any geometry. Scale the translation components by $\bar Z$, $\tilde f_0=(\bar Z f_{0,v},f_{0,\omega})$ and $\tilde p=(R^\top T/\bar Z,\sin\theta\,u)$, so that $\|\tilde p\|^2=\|T\|^2/\bar Z^2+\sin^2\theta$. For $\varepsilon\in(0,1)$ let
$N_0=\{\|T\|\le\varepsilon\bar Z,\ \theta\le\arcsin\varepsilon\}$ and $N_\pi=\{\|T\|\le\varepsilon\bar Z,\ \theta\ge\pi-\arcsin\varepsilon\}$.

\begin{theorem}[Large-weight threshold]\label{thm:large}
Let all depths be positive on $N_0$, and define
\begin{align*}
\rho_A&=\sup_{K\setminus(N_0\cup N_\pi)}\min\big\{\|\tilde f_0\|/\|\tilde p\|,\ \hat\rho^+\big\},\\
\rho_\pi&=\sup_{K\cap N_\pi}h^+\big/\sqrt{1-\varepsilon^2},\\
\rho_0&=\sup_{N_0}\lambda_{\min}(H_{\rm img})^-\big/\sqrt{1-\varepsilon^2},
\end{align*}
where $H_{\rm img}$ is the Hessian of $\tfrac12\|e\|^2$ in the chart $(T+\bar Z a,R\,\mathrm{Exp}(\omega))$. If $\rho>\bar\rho=\max\{\rho_A,\rho_\pi,\rho_0\}$, every critical point of $E_\rho$ in $K$ other than the goal has a direction of negative curvature.
\end{theorem}
\begin{proof}
A critical point $x$ at weight $\rho$ satisfies $\rho\|\tilde p\|=\|\tilde f_0\|$ and $\rho=\hat\rho(x)$, so $\rho\le\rho_A$ unless $x\in N_0\cup N_\pi$. In $N_\pi$, $\cos\theta\le-\sqrt{1-\varepsilon^2}$, and Lemma~\ref{lem:descent} applies. In $N_0$, the pose Hessian is at least $\sqrt{1-\varepsilon^2}$ in every unit direction, so $E_\rho$ is strictly convex along every geodesic of $N_0$; $N_0$ is geodesically convex (a ball times a geodesic ball of radius below $\pi/2$), so the goal is its only critical point. $N_0$ need not lie in $K$.
\end{proof}

\begin{table}[t]
\centering
\caption{Threshold of Theorem~\ref{thm:large} on the visible region $K$ (camera within $2\bar Z$ per axis, depths at least $0.25\bar Z$, image within $1.3$ times the desired extent). Sampled: constants estimated by dense sampling at the best $\varepsilon\in\{0.1,\dots,0.4\}$. Closed form: rigorous bound of Proposition~\ref{prop:closed}, at its best $\varepsilon$.}
\label{tab:large}
\small\setlength{\tabcolsep}{4pt}
\begin{tabular}{lcccccc}
\toprule
 & \multicolumn{5}{c}{Sampled} & Closed form \\
Geometry & $\varepsilon$ & $\rho_A$ & $\rho_\pi$ & $\rho_0$ & $\bar\rho$ & $\bar\rho\le$ \\
\midrule
Colotti 1a & 0.1 & 8.1 & 0 & 0.9 & 8.1 & 404 \\
Colotti 1b & 0.1 & 0.94 & 0.001 & 0.6 & 0.94 & 36 \\
Colotti 2a & 0.1 & 41 & 19 & 1.6 & 41 & 1430 \\
Colotti 2b & 0.1 & 0.87 & 0 & 0.6 & 0.87 & 48 \\
P.-E. Ex.~1 & 0.1 & 0.98 & 0 & 0.5 & 0.98 & 93 \\
\bottomrule
\end{tabular}
\end{table}

\begin{proposition}[Closed-form constants]\label{prop:closed}
Let $K$ have $n$ points with image coordinates $|x_i|,|y_i|\le b$, desired coordinates bounded by $S_0$, and depths $Z_i\ge\zeta\bar Z$. With $\eta=\sqrt2(b+S_0)$, $\Lambda_\omega^2=2b^4+2(1+b^2)^2+2b^2$ and $\Lambda^2=(1+2b^2)/\zeta^2+\Lambda_\omega^2$,
\begin{align*}
\rho_A&\le n\eta\Lambda/\varepsilon,\\
\rho_\pi&\le n\big(\Lambda_\omega^2+\eta\Lambda_\omega\sqrt{12b^2+2}\big)\big/\sqrt{1-\varepsilon^2},\\
\rho_0&\le n\min\{\eta_0,\Lambda_0 d(\varepsilon)\}S_2\big/\sqrt{1-\varepsilon^2},
\end{align*}
where $d(\varepsilon)=(\varepsilon^2+\arcsin^2\varepsilon)^{1/2}$, and $\eta_0$, $\Lambda_0$, $S_2$ are the constants above and in Appendix~B evaluated with the image bound and depth floor of $N_0$, which follow from the desired point positions.
\end{proposition}
The proof bounds the image error, the interaction matrix and the second derivatives of the features point by point (Appendix~B). We checked every bound against densely sampled poses of $K$ in each geometry.

Table~\ref{tab:large} shows the two faces of the theorem. With sampled constants, $\bar\rho$ lies between $0.87$ and $41$, inside or near the design range $0.1\le\rho\le1$ for three of the five geometries. The rigorous closed form is looser, $36$ to $1430$, but it is explicit for any geometry. Both are far above the fold, which remains the practical threshold. Near the goal the image term alone is not convex: $\lambda_{\min}(H_{\rm img})$ reaches $-0.5$ to $-1.6$ on $N_0$ at $\varepsilon=0.1$, so the pose term is needed even there.

\subsection{Box-wise certificate}
The conservatism disappears if the antiparallel condition of Lemma~\ref{lem:weight} is checked as well, since critical points lie on one-dimensional curves.
\begin{theorem}[Certificate]\label{thm:cert}
Cover $K$ by boxes $B$ in the coordinates $(T/\bar Z,\sigma)$, in the chart of Section~\ref{sec:formulation}. Suppose every box satisfies one of:
(O) some point violates a constraint of $K$ on all of $B$;
(G) $B\subset N_0$, with the convexity condition certified;
(C) $\langle f_0,p\rangle+\rho^*\|p\|^2>0$ on $B$;
(A) $1+\cos\angle(f_0,p)>0$ on $B$;
(S) for a fixed direction $v$, $v^\top(\nabla^2\tfrac12\|e\|^2+\rho\nabla^2P)v<0$ on $B$ for all $\rho\in[\max(\rho^*,\inf_B\hat\rho),\sup_B\hat\rho]$.
Then for every $\rho\ge\rho^*$, the goal is the only local minimum of $E_\rho$ in $K$.
\end{theorem}
\begin{proof}
(C) excludes critical points at weights $\rho\ge\rho^*$ by Lemma~\ref{lem:weight}; (A) excludes them at every weight; (S) shows that any critical point in $B$ has a direction of negative curvature at its own weight; (G) leaves only the goal in $N_0$.
\end{proof}

\subsection{Implementation}
In the rational chart of Section~\ref{sec:formulation}, every quantity in Theorem~\ref{thm:cert} is rational. We evaluate the tests with interval arithmetic~\cite{moore2009,tucker2011}, with outward rounding of every operation, and bound each function over a box by its value at the centre plus its interval gradient, computed by forward-mode automatic differentiation. Test (S) uses second derivatives along $v$ from interval hyper-dual numbers. The C++ implementation agrees with an independent Python model to $10^{-12}$ in all five geometries, its enclosures contained every one of $1.5\times10^6$ sampled values, and it processes about $9\times10^4$ boxes per second.

A direct search over the six pose coordinates is too costly for a global certificate near the fold: for Case 1a at $\rho^*=0.1$, interval overestimation of the Hessian (about $8\times10^3$ per unit box width in this chart) forces very small boxes near the goal, near $180^\circ$ and near the depth floor of $K$. We therefore use interval methods only locally, to certify each fold.

\subsection{Certified folds}
Interval methods do work locally. A saddle-node fold is a regular solution of the extended system
\begin{equation}\label{eq:ext}
g(y,\rho)=0,\qquad D_yg(y,\rho)\,v=0,\qquad c^\top v=1,
\end{equation}
in the 13 unknowns $(y,v,\mu)$ with $\rho=10^\mu$; regularity of~\eqref{eq:ext} at a solution is equivalent to the nondegeneracy and transversality conditions of the fold~\cite{kuznetsov2004}. We refine each numerical fold by Newton's method on~\eqref{eq:ext} and apply the Krawczyk test~\cite{moore2009}: if $K(Z)=\tilde z-C\Phi(\tilde z)+(I-CJ(Z))(Z-\tilde z)$ lies in the interior of a box $Z$, where $J(Z)$ is an interval enclosure of the Jacobian and $C\approx J(\tilde z)^{-1}$, then~\eqref{eq:ext} has exactly one solution in $Z$ and its Jacobian is nonsingular there.
\begin{proposition}[Certified folds]\label{prop:fold}
For Colotti Cases 1a, 1b, 2a and Pascual-Escudero Example 1, the branch of the IBVS local minimum of~\eqref{eq:cost} ends in a nondegenerate saddle-node fold, at a weight in
$[8.17265425, 8.17265426]\times10^{-4}$, $[1.08930976, 1.08930977]\times10^{-3}$, $[2.25064234, 2.25064235]\times10^{-4}$ and $[1.91750959, 1.91750961]\times10^{-5}$, respectively.
\end{proposition}
The Krawczyk test succeeds on boxes of radius $10^{-9}$ to $10^{-7}$ (relative) in all four cases. At a fold, the Hessian $H$ of $E_\rho$ satisfies $Hv=0$ exactly, since $H=A^\top D_yg$ at a critical point for the invertible chart map $A$. An interval $LDL^\top$ factorisation of $H\mp\epsilon I$ over the certified box, with $\epsilon$ between $1.2\times10^{-3}$ and $1.4\times10^{-2}$, gives five positive pivots for $H-\epsilon I$ and six for $H+\epsilon I$. So $H$ has exactly one zero eigenvalue and five above $\epsilon$, and the fold joins the IBVS minimum to a saddle of index 1. The enclosures hold for the exact geometry: the desired features are enclosed by interval division of the landmark coordinates rather than read as rounded constants. The proposition certifies a local event on each identified branch; that no other spurious minimum survives rests on the completeness of the numerical equilibrium search of Section~\ref{sec:maps}.

\section{Random Geometries}\label{sec:random}
To test the results beyond the five benchmarks, we generated 500 random geometries with 4, 5 or 6 points in front of the desired camera: half on a plane tilted by up to $60^\circ$, half in a 3-D cluster with a depth spread of up to $\pm50\%$. Image extents range from $0.15$ to $0.5$ (normalised), degenerate configurations are rejected, and four geometries failed numerically, leaving 496. For each geometry we followed every IBVS minimum to its fold, certified the fold as in Proposition~\ref{prop:fold}, searched for spurious minima at twice the largest fold and at $\rho=0.1$, traced and certified any minimum found there in a second pass, and evaluated the closed-form threshold of Proposition~\ref{prop:closed}.

\begin{figure*}[t]
\centering
\includegraphics[width=\textwidth]{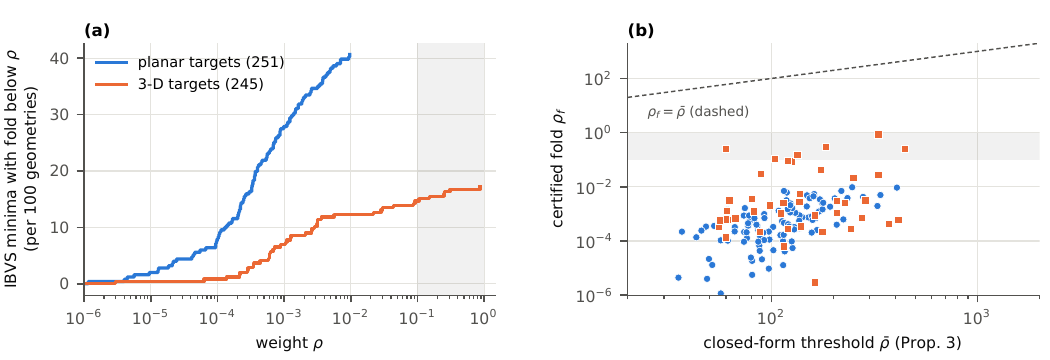}
\caption{IBVS minima in 496 random geometries. (a) Cumulative number of IBVS minima, per 100 geometries, whose fold lies below $\rho$, for planar and 3-D targets. (b) Certified fold against the closed-form threshold of Proposition~\ref{prop:closed}; the dashed line is $\rho_f=\bar\rho$. Shaded: the range $0.1\le\rho\le1$.}
\label{fig:random}
\end{figure*}

IBVS local minima are common: 143 of the 496 geometries have one (28.8\%), 40\% of planar targets and 18\% of 3-D clusters; these shares are lower bounds, since the fast search missed about one minimum in five, found later at the check weights, and a minimum whose fold lies below $10^{-4}$ can escape both checks. Every spurious minimum found at any weight continues to $\rho\to0$: none found is born at an intermediate weight. Of the 144 minima, 142 end in a fold certified by the Krawczyk test, with the desired features enclosed by interval division of the landmark coordinates, so that each certificate holds for the exact geometry.

Fig.~\ref{fig:random}(a) shows the folds. Their median is $6.3\times10^{-4}$, and 90\% lie below $9.5\times10^{-3}$. Planar targets lose every minimum below $10^{-2}$; compact 3-D clusters have a long tail, and in six geometries (1.2\%) the fold lies above $0.1$, the largest at $0.86$. Six of the 496 analysed geometries had an identified fold above $0.1$, so the design rule $\rho\ge0.1$ of~\cite{hafez2026ral} does not hold for all of them; for compact 3-D targets the fold should be computed, which takes one continuation per minimum. Fig.~\ref{fig:random}(b) checks Theorem~\ref{thm:large}: in every geometry the closed-form threshold ($32$ to $846$, median $135$) lies above the certified fold, by a factor of at least $239$.

\section{Certified Region of Attraction}\label{sec:roa}
Sections~\ref{sec:certificate} and~\ref{sec:random} locate the spurious minima. For control, one also needs a set of initial poses from which the law is guaranteed to converge. Because the law is a gradient flow, sublevel sets of $E_\rho$ are forward invariant, and the natural candidate is the largest sublevel set that is compact and free of spurious critical points.

\subsection{Escape through an object point}
Section~\ref{sec:topology} showed that sublevel sets need not be compact, because the camera centre can approach an object point $P_i$ with $E_\rho$ bounded. The flow can also move towards such a point. We placed the camera centre at distance $10^{-5}\bar Z$ from $P_i$, with $P_i$ in front of the camera, a random bearing and a random orientation, and computed the radial rate of the Gauss--Newton flow. In all five geometries and at every weight from $0.01$ to $10$, between 28\% and 37\% of these configurations move towards $P_i$, at rates up to $0.9\bar Z$ per unit time. The flow therefore approaches object points from a substantial set of nearby poses. This instantaneous test does not show that such a trajectory reaches $P_i$ in finite time, and we do not rely on it: Lemma~\ref{lem:compact} gives a level below which no trajectory can approach an object point, and $c_{\rm esc}$ is a lower bound on the cost needed to approach the boundary of $M$.
The level at which this becomes possible is
\begin{equation}\label{eq:cesc}
\begin{split}c_{\rm esc}(\rho)=\min_i\inf_{R\in\SO}\Big[&\tfrac12\sum_{j\neq i}\|s_j(P_i,R)-s_j^*\|^2\\&+\rho\Big(\frac{\|P_i\|^2}{2\bar Z^2}+1-\cos\theta\Big)\Big],\end{split}
\end{equation}
the camera centre being at $P_i$ and the term of point $i$ dropped.
\begin{lemma}[Compact sublevel sets]\label{lem:compact}
For $c<c_{\rm esc}(\rho)$ the set $S_c=\{x\in M:E_\rho(x)\le c\}$ is compact.
\end{lemma}
\begin{proof}
$S_c$ is closed in $M$ and bounded, since $E_\rho\ge\rho\|T\|^2/(2\bar Z^2)$. Let $x_k\in S_c$ converge to a point of $\partial M$, where some depth $Z_j$ vanishes. If the camera centre stays away from $P_j$, then $(X_j,Y_j)$ stays away from zero and $\|s_j\|\to\infty$, which is impossible in $S_c$. Otherwise the centre converges to some $P_i$; along a subsequence $R_k\to R$, and every term of~\eqref{eq:cost} with $j\ne i$ converges to its value at $(P_i,R)$, or diverges if $Z_j(P_i,R)=0$. Dropping the nonnegative term of point $i$ gives $\liminf E_\rho(x_k)\ge c_{\rm esc}(\rho)>c$, a contradiction. So $S_c$ is closed in the ambient space and contained in $M$.
\end{proof}
The inner problem in~\eqref{eq:cesc} has three unknowns. We bound it from below by branch and bound over the modified Rodrigues parameters $\|\sigma\|\le1$, with the interval kernel of Section~\ref{sec:certificate}. On boxes where a depth $Z_j$ is not bounded away from zero, we use $\|s_j\|\ge\sqrt{\inf(X_j^2+Y_j^2)}/\sup|Z_j|$, which holds for every positive depth. The search stops when the bound is within 1\% of the best value found, so each enclosure has relative width at most 1\%, and the upper ends agree with an independent multi-start search to four digits.

\subsection{Convergence theorem}
\begin{theorem}[Certified region of attraction]\label{thm:roa}
Let $c<c_{\rm esc}(\rho)$, and suppose the goal is the only critical point of $E_\rho$ in $S_c$. Then every trajectory of the Gauss--Newton law that starts in $S_c$ stays in $S_c$, is defined for all time, and converges to the goal.
\end{theorem}
\begin{proof}
$E_\rho$ does not increase along trajectories, so $S_c$ is forward invariant; it is compact by Lemma~\ref{lem:compact}, so solutions exist for all time. By LaSalle's invariance principle the $\omega$-limit set of a trajectory is a nonempty, connected subset of the critical points in $S_c$, which is the goal.
\end{proof}
No genericity or measure-zero argument is needed: below the lowest spurious critical value, saddles are absent too. The hypothesis is checked by a second branch and bound over six coordinates on the box $\|T_j\|\le\bar Z\sqrt{2c/\rho}$, $\|\sigma\|\le1$. A box is discarded if (L) a rigorous lower bound of $E_\rho$ on it exceeds $c$, (M) some depth is nonpositive on all of it, (N) a component of the body gradient $g=f_0+\rho p$, or $\langle g,p\rangle$, has an enclosure that excludes zero, or (G) it lies in a goal box on which the Krawczyk test proves that the goal is the only zero of $g$. Test (N) uses the first- and second-order enclosures of Section~\ref{sec:certificate}; test (L) uses the depth-robust image bound above. Unlike Theorem~\ref{thm:cert}, no second-derivative test is needed, because $S_c$ contains no saddle; this is why the six-dimensional search is feasible here. One safeguard is needed in the splitting rule: near the goal the Hessian is ill-conditioned (a translation along the image plane is nearly compensated by a rotation), and boxes split only along the dimension that dominates the enclosure width become needle-shaped and never resolve; we therefore never let a box exceed an aspect ratio of 8.

\begin{algorithm}[t]
\caption{Sublevel-set certificate (hypothesis of Theorems~\ref{thm:roa} and~\ref{thm:barrier})}\label{alg:cert}
\begin{algorithmic}[1]
\Require cost $E$ ($E_\rho$ or $E_B$), level $c$, translation bound $R_T=\sqrt{2c/\rho}$
\State Krawczyk goal box $G$ around $y=0$ with $J(0)^{-1}$ as preconditioner and the residual $g(0)$ enclosed
\State Stack $\gets$ $4^6$ boxes covering $[-R_T,R_T]^3\times[-1,1]^3$
\While{Stack not empty}
  \State pop $B$
  \If{$B$ outside $\|\sigma\|\le1$, or $B\subset G$} \textbf{continue} \EndIf
  \If{some depth (or margin) $\le0$ on all of $B$} \textbf{continue} \Comment{(M)}\EndIf
  \If{$\inf_B E>c$} \textbf{continue} \Comment{(L)}\EndIf
  \If{an enclosure of $g_k$, $\langle g,p\rangle$ or $(J(0)^{-1}g)_k$ excludes $0$} \textbf{continue} \Comment{(N)}\EndIf
  \If{$B$ below the minimum width in every coordinate} \textbf{report} $B$ unresolved; \textbf{continue} \EndIf
  \State split $B$ along the dimension that dominates the enclosure width, keeping the aspect ratio $\le8$
\EndWhile
\end{algorithmic}
\end{algorithm}

The certificate holds when no box is reported. Three details make the goal box rigorous. The desired features are not read as rounded constants: the verifier encloses $X_i^\star/Z_i^\star$ and $Y_i^\star/Z_i^\star$ by interval division of the landmark coordinates, so that $y=0$ is exactly a critical point of the model. The Krawczyk operator includes the residual term, $K(X)=-Y g(0)+(I-YJ(X))X$, with $g(0)$ enclosed rather than assumed zero. And where the goal set is a parallelotope $\{Vz:|z_k|\le r_k\}$ along the eigenvectors $V$ of $J(0)$, computed in floating point, membership of a box is tested with $V^\top$ together with a rigorous bound on $\|V^\top V-I\|_\infty$, so that the inexact orthogonality of $V$ is accounted for. The initial boxes are processed in parallel, and each finished initial box is recorded in a checkpoint file whose first line records the full configuration (geometry, weight, level, grid, goal set); a run resumes only from a checkpoint with an identical configuration. Every certificate reported below was run from a fresh checkpoint with this verifier.

\subsection{Results}
Table~\ref{tab:roa} lists $c_{\rm esc}$ and the size of $S_c$ at $c=0.9\,c_{\rm esc}^{\rm lo}$, where $c_{\rm esc}^{\rm lo}$ is the certified lower bound. In every geometry and at every weight of the grid up to $\rho=10$, the lowest spurious critical value found in Section~\ref{sec:maps} lies above $c_{\rm esc}$ (for Case 1a, $0.49$ against $0.30$ at $\rho=0.1$ and $2.37$ against $1.40$ at $\rho=1$), so the escape level, not a saddle, limits the region.

The six-dimensional search closes, with no unresolved box, for all five geometries at $\rho=1$ and for Case 1a at $\rho=0.1$ and $\rho=10$ (Table~\ref{tab:certset}). For Case 1a at $\rho=1$ and $c=1.2567=0.9\,c_{\rm esc}^{\rm lo}$ it processes $6.9\times10^8$ boxes in 2.8 hours on two cores; the Krawczyk goal box is $\|T_j\|\le2.8\times10^{-3}\bar Z$, $|\sigma_j|\le7.1\times10^{-4}$, and the residual at the goal changes neither the box nor its contraction. Before the run, the lower bound of $E_\rho$ and the gradient enclosures were checked against $3.8\times10^5$ sampled points with no violation. Hence, by Theorem~\ref{thm:roa}, every start in $S_c$ converges to the goal. A separate computation certifies that $S_c$ contains every pure rotation up to $22.69^\circ$ about any axis and every pure translation up to $0.445\bar Z$ in any direction; at $\rho=10$ these certified radii are $35.73^\circ$ and $0.667\bar Z$. The rotation ball $\{T=0,\ \theta\le r\}$ is the ball $\|\sigma\|\le\tan(r/4)$ in modified Rodrigues parameters, and the translation ball is $\{R=I,\ \|T\|\le r\bar Z\}$. Each is covered by three-dimensional boxes; a box disjoint from the ball is dropped, and on every other box the mean-value enclosure of $E_\rho$ must have its upper bound at most the exact level of the run, or the box is split. A centre point inside the ball with $E_\rho>c$ disproves the radius. The largest radius that passes is found by bisection, and the reported value is rounded down. Depths stay positive on both balls in closed form: on the translation ball $Z_i\ge Z_i^\star-r\bar Z>0$, since every radius is below the smallest $Z_i^\star/\bar Z$ (at least $0.90$), and on the rotation ball $Z_i\ge\|P_i\|\cos(\alpha_i+r)>0$, where $\alpha_i\le50.1^\circ$ is the angle of $P_i$ from the optical axis and $\alpha_i+r\le72.5^\circ$ in every case. At identical levels (all rows of Table~\ref{tab:certset} except Case 2a and Case 1a at $\rho=0.1$, which were certified at smaller levels than in Table~\ref{tab:roa}), the sampled estimates exceed the certified values by at most 3.4\%, as sampling can only overestimate a worst-direction radius; only the certified values are guarantees.

Two runs needed smaller levels. For Case 2a at $\rho=1$, the sublevel set at $0.9\,c_{\rm esc}^{\rm lo}$ reaches rotations above $100^\circ$ in some directions and comes close to an object point that the desired view sees far off-axis; we certify $c=1.0$ ($0.42\,c_{\rm esc}^{\rm lo}$), which still contains every pure rotation up to $13.93^\circ$ and every pure translation up to $0.34\bar Z$. At $\rho=0.1$ the sublevel set of Case 1a contains a long, flat valley (the camera retreats by up to $\bar Z$ while rotating by about $45^\circ$ at cost below $0.15$), and the goal is badly conditioned. With a test on the preconditioned gradient $J(0)^{-1}g$, whose components behave like the coordinates themselves near the goal, and $c=0.06$, the search closes with $7.9\times10^8$ boxes in 3.7 hours, and the certified set contains every pure rotation up to $6.70^\circ$ and every pure translation up to $0.135\bar Z$. Small weights thus pay twice: the certified region shrinks, and certifying it costs far more.

\begin{table}[t]
\caption{Escape level and certified region. $c_{\rm esc}$: certified enclosure of~\eqref{eq:cesc} (lower end shown, relative width ${\le}1\%$). Region at $c=0.9\,c_{\rm esc}^{\rm lo}$: largest pure rotation (deg) and pure translation (units of $\bar Z$) from the goal that stay in $S_c$, sampled estimate (minimum over 400 directions, which can exceed the true worst-direction radius); share of the starts of~\cite{hafez2026ral} inside $S_c$. cert.: the hypothesis of Theorem~\ref{thm:roa} is certified by the six-dimensional search (no unresolved box); Table~\ref{tab:certset} gives the certified levels and certified radii. $^a$Certified at the smaller level $c=1.0$. $^b$Certified at $c=0.06$.}
\label{tab:roa}
\centering\small
\begin{tabular}{lrrrrrc}
\hline
Geometry & $\rho$ & $c_{\rm esc}^{\rm lo}$ & rot. & transl. & starts & cert. \\
\hline
Case 1a & 0.01 & 0.159 & 10.2 & 0.21 & 2.0\% &  \\
 & 0.1 & 0.302 & 13.2 & 0.27 & 2.7\% & $\checkmark^b$ \\
 & 1 & 1.4 & 22.8 & 0.45 & 5.4\% & \checkmark \\
 & 10 & 9.23 & 36.0 & 0.69 & 8.3\% & \checkmark \\
 & 100 & 69.9 & 45.5 & 0.81 & 7.2\% &  \\
Case 1b & 0.01 & 0.27 & 18.8 & 0.32 & 6.7\% &  \\
 & 0.1 & 0.321 & 20.0 & 0.35 & 4.8\% &  \\
 & 1 & 0.81 & 28.0 & 0.47 & 2.1\% & \checkmark \\
 & 10 & 5.05 & 42.5 & 0.70 & 2.8\% &  \\
 & 100 & 44.3 & 50.5 & 0.83 & 3.1\% &  \\
Case 2a & 0.01 & 0.303 & 8.8 & 0.21 & 3.3\% &  \\
 & 0.1 & 0.498 & 10.5 & 0.27 & 4.1\% &  \\
 & 1 & 2.36 & 18.2 & 0.45 & 11.8\% & $\checkmark^a$ \\
 & 10 & 14.7 & 27.5 & 0.66 & 19.0\% &  \\
 & 100 & 115 & 35.0 & 0.83 & 17.9\% &  \\
Case 2b & 0.01 & 0.177 & 15.5 & 0.27 & 4.7\% &  \\
 & 0.1 & 0.312 & 19.8 & 0.34 & 4.5\% &  \\
 & 1 & 1.53 & 35.8 & 0.59 & 7.9\% & \checkmark \\
 & 10 & 9.84 & 53.8 & 0.83 & 10.4\% &  \\
 & 100 & 68.1 & 61.3 & 0.93 & 7.0\% &  \\
Ex.~1 & 0.01 & 0.139 & 13.8 & 0.24 & 2.3\% &  \\
 & 0.1 & 0.254 & 18.0 & 0.32 & 2.5\% &  \\
 & 1 & 1.31 & 34.0 & 0.58 & 5.8\% & \checkmark \\
 & 10 & 8.75 & 52.0 & 0.82 & 8.1\% &  \\
 & 100 & 63 & 59.3 & 0.91 & 5.8\% &  \\
\hline
\end{tabular}
\end{table}

\begin{figure}[t]
\centering
\includegraphics[width=\columnwidth]{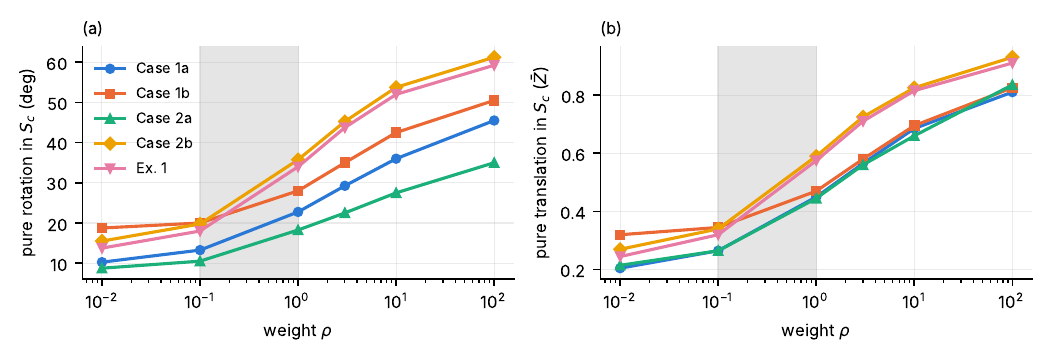}
\caption{Size of the region $S_c$ of Theorem~\ref{thm:roa} at $c=0.9\,c_{\rm esc}^{\rm lo}$ against the weight: largest pure rotation (a) and pure translation (b) from the goal that stay in $S_c$, minimum over 400 sampled directions (estimates; certified radii in Table~\ref{tab:certset}). Shaded: the range $0.1\le\rho\le1$.}
\label{fig:roa}
\end{figure}

Three trends stand out. The region grows with $\rho$ (Fig.~\ref{fig:roa}): for Case 1a from $10^\circ$ and $0.21\bar Z$ at $\rho=0.01$ to $23^\circ$ and $0.45\bar Z$ at $\rho=1$ and $46^\circ$ and $0.81\bar Z$ at $\rho=100$. It saturates, because for large $\rho$ the escape level is dominated by the pose term, $c_{\rm esc}/\rho\to\min_i\|P_i\|^2/(2\bar Z^2)$, while $E_\rho$ also grows linearly in $\rho$, so $S_c$ tends to a sublevel set of the pose term cut by the image term at the same relative level. And the certified region is conservative: it contains 2--19\% of the random starts of~\cite{hafez2026ral}, of which about 99\% converge in simulation. The certificate is limited by the escape level, which bounds a sublevel set and so covers only starts with a small initial cost, not by spurious critical points.

\begin{table}[t]
\caption{Certified sets $S_c$ (hypothesis of Theorem~\ref{thm:roa} checked by the six-dimensional search with the verifier of Section~\ref{sec:roa}, from fresh checkpoints; no unresolved box in any run). Certified radii: every pure rotation up to the angle given, about any axis, and every pure translation up to the distance given, in any direction, lies in $S_c$ (Section~\ref{sec:roa}); radii are rounded down. $c$: the exact level of the run. Boxes and wall time of the six-dimensional search alone, on two cores.}
\label{tab:certset}
\centering\footnotesize\setlength{\tabcolsep}{3pt}
\begin{tabular}{lrrrrrr}
\hline
Geometry & $\rho$ & $c$ & rot.\ ($^\circ$) & transl.\ ($\bar Z$) & boxes & time (h)\\
\hline
Case 1a & 0.1 & 0.06 & 6.70 & 0.135 & $7.9\times10^8$ & 3.7\\
 & 1 & 1.256736374 & 22.69 & 0.445 & $6.9\times10^8$ & 2.8\\
 & 10 & 8.308384781 & 35.73 & 0.667 & $1.6\times10^8$ & 0.8\\
Case 1b & 1 & 0.7293941778 & 27.84 & 0.468 & $5.4\times10^7$ & 0.3\\
Case 2a & 1 & 1.0 & 13.93 & 0.343 & $5.5\times10^8$ & 2.1\\
Case 2b & 1 & 1.375591052 & 35.61 & 0.588 & $4.0\times10^8$ & 1.4\\
Ex.~1 & 1 & 1.175520217 & 33.96 & 0.573 & $2.1\times10^8$ & 1.2\\
\hline
\end{tabular}
\end{table}

\subsection{Robustness to calibration errors}
On a robot, the law is computed from estimated features, depths and pose, and its twist passes through an estimated hand--eye transform. Write $\hat\xi(x)$ for the camera twist actually realised at pose $x$, and $\xi=-\lambda G^{-1}g$ for the ideal law. With teach-by-showing, the desired image is recorded with the same camera and the same calibration, so at the goal the estimated image error and the estimated relative pose both vanish, and $\hat\xi(x^*)=0$.
\begin{proposition}[Robust convergence]\label{prop:robust}
Let $c$ and $S_c$ satisfy the hypotheses of Theorem~\ref{thm:roa}, and let $\hat\xi$ be locally Lipschitz on $S_c$ with $\hat\xi(x^*)=0$. If the relative direction error
\[
\varepsilon(x)=\frac{\|\hat\xi(x)-\xi(x)\|_G}{\|\xi(x)\|_G},\qquad\|\zeta\|_G^2=\zeta^\top G(x)\zeta,
\]
satisfies $\varepsilon(x)<1$ for every $x\in S_c\setminus\{x^*\}$, then every trajectory of $\dot x=\hat\xi(x)$ that starts in $S_c$ stays in $S_c$ and converges to the goal. If $\varepsilon\le\bar\varepsilon<1$ on $S_c$, then $\dot E_\rho\le-\lambda(1-\bar\varepsilon)\,g^\top G^{-1}g$.
\end{proposition}
\begin{proof}
Since $g=-G\xi/\lambda$, $\dot E_\rho=g^\top\hat\xi=-\tfrac1\lambda\big(\|\xi\|_G^2+\xi^\top G(\hat\xi-\xi)\big)\le-\tfrac1\lambda(1-\varepsilon)\|\xi\|_G^2$ by the Cauchy--Schwarz inequality in the metric $G$, and $\|\xi\|_G^2=\lambda^2g^\top G^{-1}g$. So $E_\rho$ decreases strictly wherever $g\ne0$, $S_c$ is forward invariant and compact (Lemma~\ref{lem:compact}), and by LaSalle's principle every trajectory converges to the critical points in $S_c$, which by hypothesis are the goal alone.
\end{proof}
The condition is a cone condition: the realised twist must lie within an angle of the ideal twist, measured in the metric of the cost. It is the counterpart, on a certified set, of the classical sufficient condition $L\hat L^+\succ0$ for IBVS~\cite{chaumette2006visual}. Errors of the gain alone are harmless: $\hat\xi=k\xi$ gives $\varepsilon=|1-k|$, below $1$ for every $k\in(0,2)$.

We evaluated $\varepsilon$ for the calibration errors of Section~\ref{sec:experiments} (focal length $\pm3\%$, principal point $\pm5$\,px in each axis, hand--eye error of $1^\circ$ and $5$\,mm in random directions; eight combinations), with the features, depths and relative pose estimated as in the robot-in-the-loop study, by perspective-$n$-point estimation with the wrong intrinsics. On 200 points sampled from each certified set of Table~\ref{tab:certset}, plus 60 points within $10^{-2}$ to $10^{-4}$ of the goal, the largest $\varepsilon$ is $0.26$ for Case 1a at $\rho=1$ and between $0.19$ and $0.48$ for the other certified sets; the median is $0.07$ to $0.15$. Scaling all errors together, the largest $\varepsilon$ in Case 1a grows to $0.39$, $0.66$ and $1.03$ at two, four and six times the tested errors, so the condition holds up to about five times. This is sampled evidence rather than a certificate: an interval certificate would need an enclosure of the perspective-$n$-point estimate, which is the solution of an optimisation problem. Distortion errors were not modelled. This is consistent with the convergence of every certified start with calibration error in Section~\ref{sec:experiments}.

\subsection{Visibility}
Theorem~\ref{thm:roa} says nothing about features leaving the image. The sublevel set that also stays in a visible region $K$ is much smaller. With the image limited to $1.3$ times the desired extent and depths of at least $0.25\bar Z$, the largest such set, estimated by constrained multi-start optimisation, reaches only $5.5^\circ$ to $6.5^\circ$ and $0.11\bar Z$ to $0.13\bar Z$ for Case 1a and $2^\circ$ to $3.5^\circ$ for Case 1b at weights from $0.01$ to $1$; at $2.5$ times the extent, Case 1a reaches $17^\circ$ at $\rho=1$. Guaranteed visibility over large motions therefore needs a change of the potential, which we make next.

\section{Visibility by a Barrier in the Potential}\label{sec:barrier}
Section~\ref{sec:roa} certifies convergence but not visibility, and the sublevel sets of $E_\rho$ that stay visible are small. We now put the visibility constraint into the potential, so that the same gradient-flow argument gives both properties. The pose weight stays constant; the barrier acts as a state-dependent weight on the image coordinates that approach the border, in the spirit of~\cite{kermorgant2014} and of the schedules of~\cite{hafez2026whvs}, but inside a single potential.

\subsection{Barrier cost}
Let $X,Y$ be the image half-widths in normalised coordinates and $z_{\min}$ a depth floor. The margins
\begin{equation}
h_{4i-3,\dots,4i}=X\mp x_i,\ Y\mp y_i,\qquad h_{4n+i}=Z_i/\bar Z-z_{\min},
\end{equation}
define the visible set $K=\{h_j>0\ \forall j\}$. With the flat-top barrier
\begin{equation}
b(h)=\frac{(\delta-h)_+^3}{\delta^2 h}=\frac{1}{\delta^2}\Big(\frac{\delta^3}{h}-3\delta^2+3\delta h-h^2\Big)\ \ (0<h<\delta),
\end{equation}
which is $C^2$, vanishes for $h\ge\delta$ and tends to $+\infty$ as $h\to0^+$, the barrier cost is
\begin{equation}\label{eq:EB}
E_B=E_\rho+\mu\sum_j b(h_j)\quad\text{on }K .
\end{equation}
Its body gradient is $g_B=g+\mu\sum_j b'(h_j)\nabla h_j$, where $\nabla h$ is $\mp$ a row of $L$ for an image margin and $(0,0,-1,-Y_i,X_i,0)/\bar Z$ for a depth margin. The law is $v=-\lambda G^{-1}g_B$ with the metric $G$ of Section~\ref{sec:formulation}; it is a gradient flow of $E_B$ for the same reason as before. If every margin at the goal exceeds $\delta$, the barrier vanishes near the goal, which stays a nondegenerate minimum with the same Hessian.

\begin{theorem}[Visibility and convergence with a barrier]\label{thm:barrier}
Let $c>0$ and $S^B_c=\{x\in K:E_B(x)\le c\}$. (i) Every trajectory of $v=-\lambda G^{-1}g_B$ that starts in $S^B_c$ exists for all $t\ge0$ and stays in $S^B_c$, so every feature stays strictly inside the image and every depth above $z_{\min}\bar Z$. (ii) If, in addition, the goal is the only critical point of $E_B$ in $S^B_c$, every such trajectory converges to the goal.
\end{theorem}
\begin{proof}
$S^B_c$ is bounded because $E_B\ge\rho\|T\|^2/(2\bar Z^2)$, and closed in the ambient space because $E_B\to\infty$ at $\partial K$. It lies in $M$, since $z_{\min}>0$, so it is compact and no escape level is needed. Along the flow $E_B$ does not increase (Proposition~\ref{prop:flow} with $E_B$ in place of $E_\rho$), so a trajectory cannot leave $S^B_c$ and, staying in a compact set, exists for all $t\ge0$; this is (i). Part (ii) is then the LaSalle argument of Theorem~\ref{thm:roa}.
\end{proof}
The depth floor keeps the camera away from the object points, so no escape level is needed (Lemma~\ref{lem:compact}), so the level $c$ is limited only by critical points of $E_B$.

\subsection{Does the barrier create critical points?}
A barrier can create equilibria where it balances a gradient that points out of $K$. We searched for the critical points of $E_B$ with 250 random and 250 barrier-band Levenberg--Marquardt starts per setting, for Cases 1a and 2a, $\rho\in\{0.1,1\}$, $\mu\in\{0.01,0.1,1\}$ and $\delta\in\{0.05,0.15\}$ (23 of the 24 settings; one was not run), and with 400 of each for the chosen setting below in Cases 1a and 2a and Ex.~1. The barrier created no local minimum in any setting. It created at most three saddles, all at levels above the lowest saddle of $E_\rho$: for Case 1a at $\rho=1$, $3.88$ against $2.37$; for Case 2a, $4.52$ against $2.44$. In every setting the lowest spurious critical value of $E_B$ is that of $E_\rho$, the near-$180^\circ$ saddle. We use $\mu=0.1$, $\delta=0.1$ and $z_{\min}=0.25$, and evaluate the barrier sublevel set at $c=0.9$ times that value.

\subsection{Region and certificate}

\begin{figure}[t]
\centering
\includegraphics[width=\columnwidth]{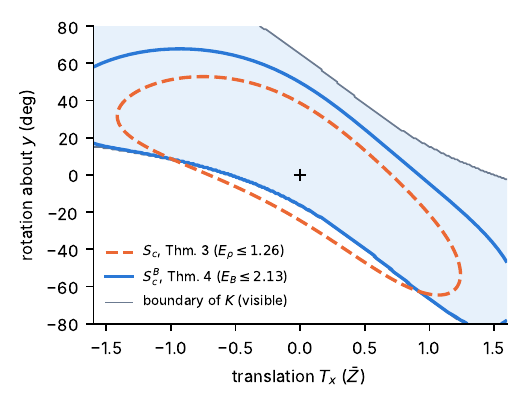}
\caption{Slice of the certified sublevel set of the smooth law and the candidate barrier sublevel set of Case 1a at $\rho=1$: camera translation along its $x$ axis against rotation about its $y$ axis, other coordinates at the goal (cross). Dashed: $S_c$ of Theorem~\ref{thm:roa} at $c=1.2567$; it extends outside the visible set $K$ (shaded). Solid: the barrier set $S^B_c$ at $c=2.132$, which stays in $K$ and reaches much further along the valley in which translation and rotation compensate in the image; its convergence relies on the numerically verified hypothesis of Theorem~\ref{thm:barrier}.}
\label{fig:slice}
\end{figure}
At these levels the barrier sublevel sets cover far more of the visible starts than the certified sets of Section~\ref{sec:roa}, with visibility guaranteed by Theorem~\ref{thm:barrier}(i); convergence from them relies on the numerically verified hypothesis of Theorem~\ref{thm:barrier}(ii). Of the starts of~\cite{hafez2026ral} that have all features in view and depths above $z_{\min}\bar Z$, $S^B_c$ contains 37\% for Case 1a at $\rho=1$ (against 14\% for the set of Section~\ref{sec:roa}), 47\% for Ex.~1 (16\%), 22\% for Case 2a (5\%) and 18\% for Case 1a at $\rho=0.1$ (0.4\%). The worst-direction radii are smaller ($10^\circ$ and $0.21\bar Z$ for Case 1a), because a pure rotation or translation soon pushes a feature to the image border; the set extends instead along the directions in which translation and rotation compensate in the image (Fig.~\ref{fig:slice}). The certifier of Section~\ref{sec:roa} carries over with the barrier added to the model: $b$, $b'$ and $b''$ are monotone on $(0,\delta]$ and zero beyond, so their enclosures follow from the end points, and a box on which some margin is nonpositive lies outside $K$. The C++ model agrees with an independent Python implementation to $10^{-12}$ at points in the barrier band, and the bounds passed $4\times10^5$ sampled checks. Wherever every margin exceeds $\delta$ on a box, $E_B=E_\rho$, and the certificate of Theorem~\ref{thm:roa} at the same or a higher level already excludes critical points; only boxes that meet the barrier band need new tests. \inprogress{six-dimensional certificate of the band: the barrier gradient varies quickly across the band and forces very small boxes; for Case 1a at $\rho=1$ and $c=1.2567$ no box has been unresolved, but boxes that contain both the goal and part of the band take hours each on two cores, against minutes without the barrier. A many-core run or a dedicated band test is needed.}

\subsection{A depth-only barrier}
The approach to object points and the visibility constraint can be separated. With the depth margins alone in~\eqref{eq:EB}, Theorem~\ref{thm:barrier} holds with $K$ replaced by $K_D=\{Z_i>z_{\min}\bar Z\ \forall i\}$: every trajectory that starts in $S^D_c=\{E_D\le c\}$ keeps every depth above $z_{\min}\bar Z$, no escape level is needed, and the trajectory converges if the goal is the only critical point of $E_D$ in $S^D_c$; the features, however, may leave the image. The band of this barrier, $Z_i<(z_{\min}+\delta)\bar Z=0.35\bar Z$, lies far from the goal. For Case 1a at $\rho=1$, a search with 600 Levenberg--Marquardt starts, half in the band, found no critical point created by the depth barrier: the lowest spurious critical value is that of $E_\rho$, $2.37$, at the saddle near $180^\circ$. At $c=0.9\times2.37=2.132$, above the escape level $1.40$, the set $S^D_c$ reaches $27.0^\circ$ and $0.52\bar Z$ in every direction (against $22.8^\circ$ and $0.45\bar Z$ for $S_c$ of Section~\ref{sec:roa}) and contains 15.4\% of the starts of~\cite{hafez2026ral} (against 5.4\%). The hypothesis is verified numerically. A six-dimensional certificate is sound for this cost and has left no box unresolved so far, but the shell of the sublevel set between the escape level and $c$ makes it long on two cores. On boxes where the barrier is inactive and $E_\rho$ is bounded by the level certified in Section~\ref{sec:roa}, the earlier certificate already excludes critical points, so only that shell needs new work.

\section{Simulated Robot-in-the-Loop Validation}\label{sec:experiments}
The theorems concern the ideal flow. We test them, and compare the hybrid laws with established laws, in a simulation that includes the effects they leave out. The laws are the smooth hybrid law at $\rho=1$ and $\rho=0.1$, the barrier law of Section~\ref{sec:barrier} at the same weights, the hybrid law of~\cite{hafez2026ral} with the $\theta u$ error at $\rho=0.1$, IBVS, PBVS, 2.5D visual servoing~\cite{malis1999} with the first point as reference, and, for Case 2a, the scheme of Kermorgant and Chaumette~\cite{kermorgant2014}.

\subsection{Setup}
A simulated 7-DOF Panda arm with the joint limits of~\cite{franka_fci} carries a camera with focal length 330\,px and image size $800\times600$ (Case 2a: $1280\times960$), radial distortion and an eye-in-hand offset. Each frame is rendered from the true camera pose with blur and pixel noise; blobs are detected and tracked, the pose is estimated by perspective-$n$-point, and the depths follow from the estimated pose. The law runs at 30\,Hz with one frame of latency, gain $\lambda=0.8$ and velocity saturation, through the arm's differential kinematics and a first-order actuator lag; the code archive submitted as multimedia contains the simulator. A run \emph{converges} when the root-mean-square feature error stays below $0.5$\,px for 1\,s; it \emph{loses a feature} when a blob is lost or comes within 4\,px of the border; it \emph{stalls} when the twist stays near zero for 3\,s with an error above 2\,px; and it ends at a \emph{joint limit} when a joint leaves its range. Each start is run with ideal calibration and with the calibration errors of~\cite{hafez2026ral}: 3\% focal length, 5\,px principal point, 20\% distortion, and $1^\circ$ and 5\,mm hand--eye error.

\subsection{Starts inside the certified region}
We drew 150 starts uniformly from the certified set $S_c$ of Case 1a at $\rho=1$ (uniform translation, Haar-distributed rotation, accepted if $E_1\le c=1.2567$). Of 175 accepted draws, 19 were unreachable by the arm and 6 had a feature outside the image: the certified set does not guarantee visibility, as Section~\ref{sec:roa} noted. The 150 starts have rotations from $7^\circ$ to $78^\circ$ (median $38^\circ$) and translations from $0.19\bar Z$ to $1.40\bar Z$; the radii of Table~\ref{tab:roa} are worst-case directions, and $S_c$ reaches much further in others.

All 300 runs of the certified law converged, with ideal and with wrong calibration (median time 7.6\,s), and so did the two other hybrid laws, PBVS and 2.5D visual servoing. IBVS converged in 92\% (ideal) and 94\% (calibration error) of the runs; the rest stalled away from the goal from starts with rotations of $37^\circ$ to $60^\circ$, or, once, hit a joint limit. No run of any law lost a feature. All 300 trials converged under the tested sensing and actuation conditions, effects that the continuous-time theorem does not model, and the certified set is not trivial: it contains starts from which IBVS fails.

\subsection{Wide start sets}
Four further sets probe the laws beyond the certified regions. For Case 1a, 150 starts come from the start box of~\cite{hafez2026ral} (E2: translations in $[-1.5,1.5]^2\times[-1.5,0.3]\,\bar Z$, rotations up to 2.2\,rad), and 150 have the same translations with Haar-distributed rotations (E3: $21^\circ$ to $178^\circ$, median $98^\circ$). For Case 2a, whose desired features sit far off-axis (normalised $y$ down to $-1.19$ with the image edge at $-1.45$), 100 starts of each kind form E4 and E5. Table~\ref{tab:sil} summarises Case 1a and Table~\ref{tab:border} the visibility in E3--E5; a script in the code archive regenerates every count from the per-run records.

Every run of the smooth hybrid and barrier laws either converged or reached a joint limit; the earlier $\theta u$ hybrid law also lost a feature in two runs. Joint limits dominate at large rotations (21 to 29 runs of each law other than IBVS in E3, almost all from starts above $100^\circ$), because the Panda wrist cannot follow the rotation the camera has to make; they are outside the present guarantees. On the starts from which no law hit a joint limit, the three hybrid laws converged in every run of every set, and so did the scheme of~\cite{kermorgant2014} in E4 and E5. IBVS stalled in 8\% to 20\% of the runs of each set. The saddles near $180^\circ$ (Proposition~\ref{prop:morse}) did not trap any run, as expected of saddles.

\subsection{Where the guarantee matters}
Success rates do not separate the hybrid laws from PBVS, 2.5D visual servoing or the scheme of~\cite{kermorgant2014}: all converge almost everywhere once joint limits are excluded. What separates them is visibility. Table~\ref{tab:border} counts, over the wide sets E3--E5, whose start poses are identical for all laws, the runs that lost a feature and those whose recorded margin to the image border fell below 20\,px, with the joint-limit stops alongside. PBVS lost a feature in 24 runs and had a recorded margin below 20\,px in 38; IBVS lost a feature in 31 runs (38 with a lost feature or a margin below 20\,px), 2.5D visual servoing in 4 (13) and the $\theta u$ hybrid law in 2 (12). The smooth hybrid and barrier laws lost none. Their smallest margins, 10.2\,px in E3, 19.7\,px in E4 and 10.9\,px in E5, are the same for every law that kept all features in view, which indicates that the starts, not the laws, set them. For the barrier law, Theorem~\ref{thm:barrier} turns this observation into a guarantee on $S^B_c$; for PBVS, 2.5D visual servoing and the scheme of~\cite{kermorgant2014}, no such statement is available.

\begin{table}[t]
\caption{Visibility in the wide sets E3--E5 (matched starts, both calibrations). JL: runs ended at a joint limit, excluded from the other columns. FL: runs that lost a feature. ${<}20$: runs whose recorded minimum margin to the image border fell below 20\,px. The simulator can stop on tracking failure before it logs that frame's margin, so the union is given separately. $^*$E4 and E5 only.}
\label{tab:border}
\centering\footnotesize\setlength{\tabcolsep}{3.5pt}
\begin{tabular}{lrrrrr}
\hline
Law & JL & runs & FL & ${<}20$\,px & FL or ${<}20$\\
\hline
Smooth hybrid, $\rho=1$ & 80 & 620 & 0 & 10 & 10\\
Smooth hybrid, $\rho=0.1$ & 78 & 622 & 0 & 10 & 10\\
Barrier, $\rho=1$ & 81 & 619 & 0 & 10 & 10\\
Barrier, $\rho=0.1$ & 82 & 618 & 0 & 10 & 10\\
Hybrid $\theta u$~\cite{hafez2026ral}, $\rho=0.1$ & 72 & 628 & 2 & 10 & 12\\
2.5D VS~\cite{malis1999} & 71 & 629 & 4 & 11 & 13\\
Kermorgant--Chaumette~\cite{kermorgant2014}$^*$ & 24 & 376 & 0 & 4 & 4\\
PBVS & 66 & 634 & 24 & 38 & 38\\
IBVS & 176 & 524 & 31 & 19 & 38\\
\hline
\end{tabular}
\end{table}

\subsection{The barrier law}
We ran the barrier law at $\rho=1$ and $\rho=0.1$ on E3, E4 and E5, 1400 runs with ideal and with wrong calibration. No run lost a feature or stalled; every run converged or ended at a joint limit, and the counts match those of the law without barrier to within three runs. In these tests the hybrid law already keeps the features in view, so the barrier is rarely active; its value is that visibility is now guaranteed from the whole of $S^B_c$ rather than observed.

\begin{figure*}[t]
\centering
\includegraphics[width=\textwidth]{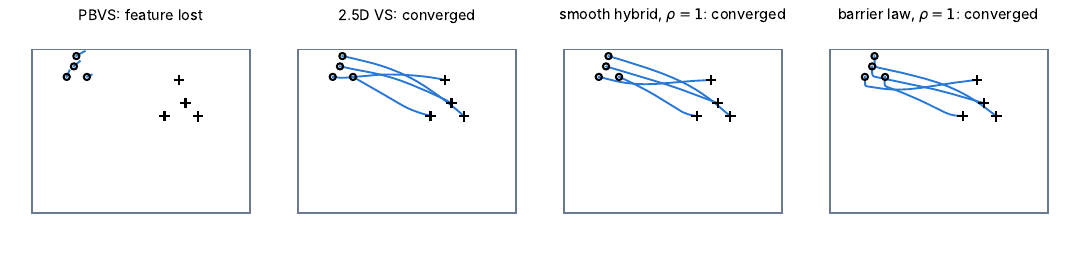}
\caption{Image trajectories from one start of the wide set E3 (Case 1a, rotation $109^\circ$, ideal calibration). Circles: initial features; crosses: desired features; frame: image border. PBVS moves the features out of the image; 2.5D visual servoing, the smooth hybrid law and the barrier law converge.}
\label{fig:traj}
\end{figure*}

\paragraph*{Terminal pose} Convergence is declared on the image error, but the records also hold the true terminal pose error. Every converged run of the smooth hybrid law ended within $1.76$\,mm and $0.48^\circ$ of the desired pose, and every converged run of the barrier law within $1.41$\,mm and $0.34^\circ$ (largest values over all sets and both calibrations, with $\bar Z=0.2$\,m). The corresponding maxima are $1.77$\,mm and $0.48^\circ$ for the $\theta u$ hybrid law, $2.37$\,mm and $0.58^\circ$ for 2.5D visual servoing, $2.28$\,mm and $0.58^\circ$ for PBVS, $2.65$\,mm and $0.68^\circ$ for IBVS, and $0.75$\,mm and $0.19^\circ$ for the scheme of~\cite{kermorgant2014} on Case 2a. A converged run therefore never ended at a wrong pose with the right image.

Taken together, the experiments place the hybrid law with the best of the established laws in practice: in all 2600 runs of the smooth hybrid law (both weights, five sets) it never stalled and never lost a feature, 2438 runs converged and the other 162 ended at a joint limit; the 1400 runs of the barrier law behaved the same (Fig.~\ref{fig:traj}). What separates it from them is not the success rate in these tests but the guarantee: for the hybrid law the convergence from $S_c$ is a theorem, checked by a computer-assisted proof, whereas for PBVS, 2.5D visual servoing and the scheme of~\cite{kermorgant2014} it is an observation.

\begin{table}[t]
\caption{Simulated robot-in-the-loop outcomes, Case 1a, 150 starts per set; each entry ideal/wrong calibration. Converged runs for the certified set $S_c$ (E1), the start box of~\cite{hafez2026ral} (E2) and the wide set with Haar rotations (E3); for E3 also runs that lost a feature (FL) or hit a joint limit (JL). Runs not counted as converged, lost features or joint limits stalled away from the goal (IBVS only).}
\label{tab:sil}
\centering\footnotesize\setlength{\tabcolsep}{3pt}
\begin{tabular}{lrrrrr}
\hline
Law & E1 & E2 & E3 & E3 FL & E3 JL \\
\hline
Smooth hybrid, $\rho=1$ & 150/150 & 149/149 & 121/121 & 0/0 & 29/29 \\
Smooth hybrid, $\rho=0.1$ & 150/150 & 149/149 & 123/123 & 0/0 & 27/27 \\
Hybrid $\theta u$~\cite{hafez2026ral}, $\rho=0.1$ & 150/150 & 149/149 & 125/125 & 1/1 & 24/24 \\
PBVS & 150/150 & 149/149 & 122/122 & 7/7 & 21/21 \\
2.5D VS~\cite{malis1999} & 150/150 & 148/148 & 125/125 & 2/2 & 23/23 \\
IBVS & 138/141 & 126/125 & 70/73 & 10/10 & 41/37 \\
\hline
\end{tabular}
\end{table}

\section{Discussion}\label{sec:discussion}
\begin{table}[t]
\caption{Status of the results. Proved: for any geometry. Certified: computer-assisted proof by interval arithmetic for the geometries listed. Numerical: evidence, not proof.}
\label{tab:status}
\centering\scriptsize\setlength{\tabcolsep}{3pt}
\begin{tabular}{p{3.45cm}p{2.4cm}l}
\hline
Result & Scope & Status\\
\hline
Lemmas~\ref{lem:domain}, \ref{lem:rot}; Props.~\ref{prop:flow}, \ref{prop:goal} (structure, gradient flow, goal) & any geometry, $\rho>0$ & proved\\
Prop.~\ref{prop:morse} (Morse count) & under Assumption~\ref{as:proper}, which fails on the benchmarks & proved, conditional\\
Thm.~\ref{thm:large}, Prop.~\ref{prop:closed} (explicit threshold) & any geometry & proved\\
Thm.~\ref{thm:cert} (box-wise threshold) & any geometry & proved; not run globally\\
Folds of the IBVS minima (Prop.~\ref{prop:fold}) & 4 benchmarks, 142 random minima & certified\\
Lemma~\ref{lem:compact}, Thm.~\ref{thm:roa} (region of attraction) & any geometry & proved\\
Escape level $c_{\rm esc}$ & 5 geometries, 5 weights & certified\\
Hypothesis of Thm.~\ref{thm:roa} & 5 geometries at $\rho=1$; Case 1a at $\rho=0.1,10$ & certified\\
Prop.~\ref{prop:robust} (calibration errors) & any geometry & proved\\
Cone condition of Prop.~\ref{prop:robust} & 7 certified sets, tested errors & numerical\\
Thm.~\ref{thm:barrier}(i) visibility, (ii) convergence & any geometry & proved\\
Hypothesis of Thm.~\ref{thm:barrier}(ii) & Cases 1a, 2a, Ex.~1 & numerical\\
Hypothesis with depth-only barrier & Case 1a, $\rho=1$ & numerical\\
Critical-point maps, Morse test, random geometries & 5 benchmarks, 496 random & numerical\\
Robot-in-the-loop study & Cases 1a, 2a & simulation\\
\hline
\end{tabular}
\end{table}
\paragraph*{What is guaranteed} Three kinds of statements appear in this paper, and they should not be confused. Table~\ref{tab:status} lists them. Theorems~\ref{thm:large},~\ref{thm:roa} and~\ref{thm:barrier} and Lemma~\ref{lem:compact} are proved for any geometry. Their hypotheses are checked for specific geometries by computer-assisted proofs: the certified folds (Proposition~\ref{prop:fold}), the escape levels, and the six-dimensional sublevel-set searches, which rely on interval arithmetic with outward rounding and on the floating-point assumptions of Appendix~C. The critical-point maps, the Morse test, the random-geometry study and the barrier critical-point search are numerical evidence, not proofs; they choose the levels that the certificates then check, and they explain why the fold, not the explicit threshold, is the practical threshold.

\paragraph*{Why the certified regions are smaller than the observed basin} About 99\% of the random starts of~\cite{hafez2026ral} converge in simulation, but the certified sets contain only some of them. The sets are sublevel sets: a start with a large image error has a large cost even when its trajectory would converge. Without a barrier, the level is further capped by the escape level of the object points, which lies below the lowest spurious saddle in every geometry. The barrier removes that cap and enlarges the set substantially (Section~\ref{sec:barrier}), although there the absence of further critical points is so far verified numerically, not certified, but a sublevel set of a cost that measures distance to the goal will never cover starts that are far in cost and yet converge. Certificates on non-sublevel sets, for example by reachability or by the stability-boundary theory of~\cite{chiang1988}, would be the next step.

\paragraph*{The weight} Small weights pay twice. Below the fold the IBVS minimum survives; just above it, the certified sets are small and their certificates are expensive, because the cost has a long flat valley in which a retreat of the camera is compensated by a rotation, and the goal is badly conditioned (Case 1a at $\rho=0.1$ needed $7.9\times10^8$ boxes for a set whose certified rotation radius is $6.70^\circ$). Large weights give larger certified sets ($35.73^\circ$ at $\rho=10$) but weaken the image term, which is the term that keeps the features in view without a barrier~\cite{hafez2026ral}. With the barrier, visibility no longer depends on the image term, and the weight can be chosen for the size of the certified set. Six of the 496 analysed random geometries had an identified fold above $0.1$, so the range $0.1\le\rho\le1$ of~\cite{hafez2026ral} does not always suffice; compact 3-D clusters can need more, and their fold takes one continuation per IBVS minimum to compute.

\paragraph*{Limitations} The guarantees concern the continuous-time gradient flow. Proposition~\ref{prop:robust} extends them to estimated features, depths and pose under a cone condition that we checked by sampling, not certified. The simulated robot-in-the-loop study shows that sampling, latency, saturation and image noise do not change the outcome, but it does not prove it; the sampled-data certificates of~\cite{hafez2026whvs} address that gap for weight schedules. The model uses point features and a static target, and the certificates are per geometry: each new geometry needs its own run, from minutes to about a day on two cores. Joint limits, not the law, ended most failed runs at large rotations; they are outside the present guarantees. The topology of the pose domain precludes global asymptotic stabilisation by any continuous autonomous feedback law~\cite{bhat2000}; this does not determine where the remaining critical points lie, and the certified sets exclude every critical point other than the goal. Finally, the barrier was rarely active in our tests, because the hybrid law already keeps the features in view; its value is the guarantee, not a higher success rate.

\section{Conclusion}\label{sec:conclusion}
The hybrid law with a smooth rotation error is a gradient flow whose goal, for topological reasons, cannot be globally attractive, and whose spurious critical points found in five benchmark geometries are, above a small weight, all saddles near $180^\circ$. We proved an explicit weight threshold for any geometry, certified the folds that end the IBVS minima, bounded the cost needed to approach an object point by a certified escape level, and proved convergence on sublevel sets whose hypotheses a six-dimensional computer-assisted proof checks; a barrier inside the potential adds visibility and removes the escape, under a hypothesis that we verified numerically. In extensive simulated robot-in-the-loop tests the hybrid law performs as well as the best established laws, and unlike them it carries certified guarantees of convergence from explicit regions. Certifying the barrier hypothesis and the calibration cone condition, extending the certificates to sampled-data execution, and enlarging them beyond sublevel sets are the natural next steps.

\section*{Disclosure of AI use}
An AI assistant (Claude, Anthropic) was used in preparing this manuscript: drafting and editing of text in all sections, writing and running the analysis, certification and simulation code, and producing the figures. The author verified the mathematics, the code, the results and the references, and takes full responsibility for the content.

\appendix
\section*{Appendix A: Proof of Lemma~\ref{lem:rot}(iii)}
For $f(R)=\tfrac12(3-\operatorname{tr}R)$ and $R=\mathrm{Exp}(\theta u)$, $\frac{d^2}{dt^2}f(R\,\mathrm{Exp}(t\omega))\big|_{t=0}=-\tfrac12\operatorname{tr}(R[\omega]_\times^2)=\tfrac12\big((1+\cos\theta)\|\omega\|^2-(1-\cos\theta)(u^\top\omega)^2\big)$, using $[\omega]_\times^2=\omega\omega^\top-\|\omega\|^2I$ and $\operatorname{tr}R=1+2\cos\theta$. We checked the formula numerically at 200 random rotations (largest deviation $2\times10^{-8}$).

\section*{Appendix B: Closed-form constants}
For a point with normalised coordinates $|x|,|y|\le b$ and depth $Z\ge\zeta\bar Z$, the translation columns of $L$, scaled by $\bar Z$, have norm at most $\sqrt{1+2b^2}/\zeta$ (the $2\times3$ matrix $[-I_2\ s]/Z$ has singular values $1/Z$ and $\sqrt{1+\|s\|^2}/Z$), and the rotation columns have Frobenius norm at most $\Lambda_\omega$. Hence $\|\tilde f_0\|\le\sum_i\|\tilde L_i\|\|e_i\|\le n\Lambda\eta$, and $\rho_A\le n\eta\Lambda/\varepsilon$ because $\|\tilde p\|\ge\varepsilon$ outside $N_0\cup N_\pi$.

Along $t\mapsto(T,R\,\mathrm{Exp}(tu))$ the features obey $\dot s=L_\omega(s)u$ and $\ddot s=\sum_k\partial_kL_\omega(s)u\,\dot s_k$, with $\|\partial_xL_\omega\|_F^2,\|\partial_yL_\omega\|_F^2\le6b^2+1$, which gives $h\le\sum_i(\|\dot s_i\|^2+\|e_i\|\|\ddot s_i\|)\le n(\Lambda_\omega^2+\eta\Lambda_\omega\sqrt{12b^2+2})$.

Along a unit chart curve $t\mapsto(T+t\bar Za,R\,\mathrm{Exp}(t\omega))$, the body velocity $\xi=(R^\top a,\omega)$ has $\|\xi\|=1$ and $\|\dot\xi\|\le1$, the depth rate satisfies $|\dot Z|/Z\le1/\zeta+\sqrt2 b$, and
\[
\|\ddot s_i\|\le S_2=\Lambda\sqrt{2/\zeta^2+12b^2+2}+\frac{\sqrt{1+2b^2}}{\zeta}\Big(\frac1\zeta+\sqrt2b\Big)+\Lambda .
\]
Since $H_{\rm img}=\sum_i(J_i^\top J_i+e_i^\top\nabla^2s_i)$, $\lambda_{\min}(H_{\rm img})\ge-\sum_i\|e_i\|S_2$. On $N_0$, $e_i$ vanishes at the goal and grows at most at rate $\Lambda_0$ along a geodesic of length at most $d(\varepsilon)$, so $\|e_i\|\le\min\{\eta_0,\Lambda_0d(\varepsilon)\}$. The image bound $b_0$ and depth floor $\zeta_0$ on $N_0$ follow from $\|R^\top(P_i-T)-P_i\|\le\varepsilon\bar Z+\arcsin(\varepsilon)\|P_i\|$.

\section*{Appendix C: Floating-point assumptions}
All certified computations (Proposition~\ref{prop:fold} and the box tests of Theorem~\ref{thm:cert}) use interval arithmetic in IEEE double precision with round-to-nearest. Every interval operation widens its result outward by $2^{-51}$ relative plus the smallest normal number, which dominates the rounding error of one operation. The only transcendental functions are \texttt{exp} and \texttt{log} (for $\rho=10^\mu$); we assume the C library computes them to within one unit in the last place and widen their results by $10^{-15}$ relative. Rotations use modified Rodrigues parameters, so the model itself needs only $+,-,\times,\div$.

\bibliographystyle{IEEEtran-doi}
\bibliography{refs}
\end{document}